\documentclass[conference,compsoc]{IEEEtran}

\AtBeginDocument{%
  }

\usepackage{listings}
\usepackage{bbding}
\usepackage{tikz}
\usepackage{amsmath}
\usepackage{wasysym}
\usepackage{tablefootnote}
\usepackage{tikz}
\def\checkmark{\tikz\fill[scale=0.4](0,.35) -- (.25,0) -- (1,.7) -- (.25,.15) -- cycle;} 

\usepackage{filecontents}

\let\oldthefootnote\thefootnote

\renewcommand{\thefootnote}{\fnsymbol{footnote}}

\usepackage{lscape}
\usepackage{multirow}
\usepackage{graphicx}
\usepackage{booktabs}
\usepackage{soul}
\usepackage{algorithm}
\usepackage{algpseudocode}
\usepackage{amsmath}
\usepackage{graphicx}

\usepackage{bm}

\usepackage{xcolor}
\usepackage{xspace}

\usepackage{colortbl}
\usepackage{booktabs} 

\usepackage{subcaption}

\PassOptionsToPackage{breaklinks = true,unicode = true,urlcolor = blue,colorlinks = true,citecolor = red,linkcolor = red}{hyperref}
\usepackage{hyperref}

\usepackage{multirow}
\usepackage{array}
\newcolumntype{P}[1]{>{\centering\arraybackslash}p{#1}} 

\usepackage{enumitem}

\usepackage{multicol}
\usepackage{multirow}

\usepackage{tabularx}
\usepackage{balance}
\usepackage{epstopdf}

\usepackage{amsfonts}
\usepackage{amsthm}
\usepackage{amsmath}

\usepackage[nameinlink,capitalize]{cleveref} 
\crefname{appendix}{Appendix}{Appendices}
\Crefname{appendix}{Appendix}{Appendices}

\usepackage[font=small]{caption}

\newcommand{\paragraphb}[1]{\vspace{0.75ex}\noindent{\bf #1.} }
\newcommand{\paragraphbe}[1]{\vspace{0.75ex}\noindent{\bf \em #1} }

\newcommand{\paragraphq}[1]{\vspace{0.75ex}\noindent{\bf #1} }

\newcommand{\ignore}[1]{}

\definecolor{orange}{RGB}{255,127,80}
\definecolor{darkgreen}{RGB}{50,127,0}
\definecolor{Blue}{RGB}{0,0,255}

\usepackage{comment}
\usepackage{amsmath}

\newcommand{\sgd}{SGD}
\newcommand{\dpsgd}{DP-SGD}

\newcommand{\cifar}{CIFAR-10}

\newcommand{\pr}[1]{\ensuremath{\mathrm{Pr}(#1)}} 

\newcommand{\dpalg}{\ensuremath{F}}

\newtheorem{definition}{\bf Definition}
\newtheorem{lemma}{\bf Lemma}

\newcounter{mynote}[section]

\newcommand{\thenote}{\thesection.\arabic{mynote}}

\newcommand{\wbnote}[1]{\ifx\outforreview\undefined\refstepcounter{mynote}{\it\textcolor{blue}{(WB~\thenote: { #1})}}\fi}

\newcommand{\vbnote}[1]{\ifx\outforreview\undefined\refstepcounter{mynote}{\it\textcolor{darkgreen}{(VB~\thenote: { #1})}}\fi}

\newcommand{\fixme}[1]{\ifx\outforreview\undefined\textbf{\textcolor{red}{[FIXME: #1]}}\fi}

\usepackage{tcolorbox} 
\usepackage{changepage}

\theoremstyle{break}
\newtheorem{takeawaythm}{Takeaway}

\newcommand{\takeaway}[1]{\begin{tcolorbox}[boxrule=1pt,arc=0.3em,boxsep=-1mm,width=0.975\linewidth, colback=white!97!black]\begin{takeawaythm}#1\end{takeawaythm}\end{tcolorbox}}
\usepackage{comment}
\newcommand{\selfmix}{\ensuremath{\textsc{DP-Mix}_{\textsc{Self}}}\xspace} 
\newcommand{\diffmix}{\ensuremath{\textsc{DP-Mix}_{\textsc{Diff}}}\xspace}

\usepackage{balance}

\usepackage{titlesec} 

\begin{document}


\title{Towards Reliable and Generalizable Differentially Private Machine Learning (Extended Version)}


\author{\IEEEauthorblockN{Wenxuan Bao}
\IEEEauthorblockA{\textit{University of Florida} \\
Gainesville, USA \\
wenxuanbao@ufl.edu}
\and
\IEEEauthorblockN{Vincent Bindschaedler}
\IEEEauthorblockA{\textit{University of Florida} \\
Gainesville, USA \\
vbindschaedler@ufl.edu}}
\maketitle

\begin{abstract}
  There is a flurry of recent research papers proposing novel differentially private machine learning (DPML) techniques. These papers claim to achieve new state-of-the-art (SoTA) results and offer empirical results as validation. However, there is no consensus on which techniques are most effective or if they genuinely meet their stated claims. Complicating matters, heterogeneity in codebases, datasets, methodologies, and model architectures make direct comparisons of different approaches challenging.

In this paper, we conduct a reproducibility and replicability (R+R) experiment on 11 different SoTA DPML techniques from the recent research literature. Results of our investigation are varied: while some methods stand up to scrutiny, others falter when tested outside their initial experimental conditions.  We also discuss challenges unique to the reproducibility of DPML, including additional randomness due to DP noise, and how to address them. Finally, we derive insights and best practices to obtain scientifically valid and reliable results. 
\end{abstract}

\begin{IEEEkeywords}
Differential Privacy, Machine Learning, Reproducibility
\end{IEEEkeywords}



\section{Introduction\protect\footnotemark}
\footnotetext{This paper is published at ACSAC 2024. This is the extended version that includes an overview of the relevant literature. We open-source our codebase at: \url{https://github.com/wenxuan-Bao/Reliable-and-Generalizable-DPML}.}

\renewcommand{\thefootnote}{\oldthefootnote}
\setcounter{footnote}{0} 
The reproducibility crisis that plagues machine learning (ML) and ML-based science is well-documented~\cite{kapoor2023leakage,pineau2021improving,semmelrock2023reproducibility}. 
Due to the breadth and diversity of machine learning applications, studies are conducted within specific domains and application areas such as healthcare~\cite{beam2020challenges,mcdermott2021reproducibility}, life sciences~\cite{heil2021reproducibility}, and security~\cite{olszewski2023get,daoudi2021lessons}. 

The reproducibility of differentially private machine learning (DPML) has so far received little attention. In a nutshell, DPML seeks to protect the privacy of training data of a model using the mathematical framework of differential privacy (DP)~\cite{dwork2006calibrating,dwork2006our,dwork2014algorithmic}. 
While machine learning models are typically trained with Stochastic Gradient Descent~\cite{gower2019sgd} (SGD), DPML primarily leverages DP-SGD~\cite{abadi2016deep}, a drop-in replacement for SGD. DP-SGD iteratively updates model parameters using the gradient like SGD, but also clips individual training points' gradients prior to aggregation, and adds Gaussian noise to updates. DP-SGD provably satisfies differential privacy, but the model predictions quality is often drastically degraded.

Recently, numerous research papers proposing novel DPML techniques claim to achieve new state-of-the-art (SoTA) results~\cite{klause2022differentially,sander2022tan,dormann2021not,de2022unlocking,bao2024dp,tramer2020differentially,bu2022scalable,bu2022automatic,cattan2022fine,luo2021scalable,tang2023differentially}. 
These techniques and the DPML literature more broadly have delivered remarkable improvements in the utility-privacy tradeoff since DP-SGD was introduced by Abadi et al~\cite{abadi2016deep} in 2016. Despite this, there is little consensus regarding which techniques are most effective. 
This is in part the result of the heterogeneity in codebases, datasets and model architecture, making apples-to-apples comparisons difficult to obtain.

In this paper, we systematically survey, taxonomize, and discuss the research literature on differentially private machine learning (DPML). Our goal is to explicate the various competing approaches that have been proposed and identify promising directions for future work.

Then we conduct a reproducibility and replicability (R+R) experiment on 11 different recent DPML techniques. We focus on centralized machine learning --- leaving federated learning for future work --- and computer vision tasks, since these have received significant recent attention and show great promise. 

However, the purpose of our investigation is not to point fingers or cast any specific work in a negative light. We seek to understand what methodological steps make DPML research reproducible and replicable and lead to reliable findings.
We discover that a significant challenge with DPML reproducibility --- as opposed to (non-private) ML reproducibility --- is the additional randomness (e.g., noise added to gradients, etc.). Variability in measured results is often substantial, especially when few runs are performed and (or) when few datasets/models are used. Averaging results of multiple runs could alleviate this issue. But DPML training is much slower than non-private ML training~\cite{bao2022importance,subramani2021enabling,bu2022scalable} so performing extensive evaluation is a major computational burden. 

The net effect of this additional randomness is that reliable results are more difficult to obtain. A single lucky run with higher performance than the baseline may be (wrongly) interpreted as a new SoTA result.

To overcome this, we propose a framework based on paired t-tests~\cite{ross2017paired} and Cohen's d~\cite{cohen2013statistical}. This framework allows us to determine which of our selected techniques indeed outperform their baselines and also quantifies the additional variability of DPML compared to non-private ML.

Stepping back, the concrete goals of our investigation are threefold: (1) quantify the reproducibility of existing work to confirm or disconfirm SoTA claims, thereby separating the wheat from the chaff; (2) identify which techniques provide improvements that are scientifically sound and generalize beyond the (necessarily) narrow experimental setting of their originating papers; and (3) establish guidelines and best practices that future research can adopt to maximize fair comparisons and reduce false discovery risk.

The results of our investigations are mixed. Obtaining implementations of the selected techniques was not a problem. Codebases were readily available in many cases and when they were not, techniques could easily be implemented based on research papers' descriptions.  We were able to directly obtain results consistent with what all 11 selected papers reported. However, when using the techniques outside of the narrow experimental settings of their original papers (e.g., on a different dataset or with a different model architecture), only 7 out of 11 completely delivered their claimed improvements.

Among those techniques that disappointed in new experimental settings, we found notable methodological pitfalls such as a lack of ablation studies, narrow sets of evaluation tasks, and results reported for a single run only.

To avoid these pitfalls in future research and maximize the chance of reproducibility and replicability, we derive guidelines and a concrete checklist.

\paragraphq{Summary of contributions:}
\begin{itemize}[leftmargin=1em]

\item In this paper, we systematically survey, taxonomize, and discuss the research literature on DPML. We show our taxonomy of DPML in~\cref{tab:framework} 

\item We conduct a thorough reproducibility and replicability evaluation of 11 recent SoTA DPML techniques, showing their variance in generalizability and reliability. 

\item We introduce a framework utilizing paired t-tests, specifically tailored to assess the inherent variability and reproducibility challenges unique to DPML.

\item We propose comprehensive guidelines and a checklist to enhance future DPML research, aiming to standardize practices and reduce the prevalence of invalid claims. 

\end{itemize}

\begin{table*}[h]
\caption{Taxonomy of Methods.}
\label{tab:framework}
\resizebox{\textwidth}{!}{%
\begin{tabular}{lll}
\toprule
Category                           & References & Section \\ \midrule
Tuning model architecture  & \cite{bao2022importance} , \cite{priyanshu2021efficient}, \cite{cheng2022dpnas}, \cite{remerscheid2022smoothnets},\cite{wang2021dplis}, \cite{shamsabadi2021losing}, \cite{papernot2021tempered},\cite{klause2022differentially}, \cite{sander2022tan}            & \cref{sec:experiment_direction:model}        \\
Feature selection          &  \cite{abadi2016deep}, \cite{tramer2020differentially}, \cite{bao2022importance}, \cite{singhal2021privately}         & \cref{sec:experiment_direction:train}        \\
 Hyperparameter tuning      & \cite{bao2022importance}, \cite{de2022unlocking}, \cite{sander2022tan} ,\cite{papernot2021hyperparameter}, \cite{dormann2021not}, \cite{wang2023dp},\cite{bao2024dp},\cite{park2024distribution}          &   \cref{{sec:hyperparam}}      \\
Gradient clipping          &  \cite{abadi2016deep}, \cite{chen2020understanding}, \cite{lee2021scaling}, \cite{bu2021fast}, \cite{bu2022scalable}, \cite{yu2018improve}, \cite{pichapati2019adaclip}, \cite{andrew2021differentially}, \cite{du2021dynamic}, \cite{lin2022understanding}, \cite{golatkar2022mixed},  \cite{bu2022automatic}, \cite{yang2022normalized},  \cite{mcmahan2018learning}, \cite{van2018three},\cite{liu2021differentially}, \cite{xu2020removing}, \cite{bu2021convergence}, \cite{xia2023differentially}, \cite{koloskova2023revisiting}, \cite{xiao2023geometry}, \cite{xiao2023theory}          &  \cref{sec:clipping}       \\

Transfer learning          & \cite{mehta2022large}, \cite{cattan2022fine}, \cite{bu2022differentiallyb}, \cite{golatkar2022mixed},\cite{amid2022public},  \cite{luo2021scalable},  \cite{hoory2021learning},  \cite{li2021large},  \cite{yu2021differentially}, \cite{tramer2022considerations}, \cite{lowy2023optimal}, \cite{tang2023differentially}, \cite{yu2023selective}, \cite{mireshghallah2022differentially},\cite{wang2024dpadapter}          & \cref{sec:public_data}          \\\midrule
Sparse Training    & \cite{zhu2021differentially}, \cite{kasiviswanathan2021sgd},  \cite{sidahmed2021efficient},  \cite{bao2022importance}, \cite{yu2021large}, \cite{ito2022scaling}, \cite{zhang2021wide}, \cite{asi2021private}, \cite{nasr2020improving}, \cite{zhou2020bypassing}, \cite{yu2020not} ,\cite{gu2023choosing}, \cite{feng2023spectral}           & \cref{sec:misc:sparse}         \\ 
Better privacy accounting & \cite{asoodeh2020better} ,  \cite{harremoes2011pairs},\cite{koskela2020computing}, \cite{koskela2021tight}, \cite{koskela2021computing}, \cite{gopi2021numerical},\cite{koskela2021tight}, \cite{chourasia2021differential}, \cite{ye2022differentially},  &
  \cref{sec:theory:account} \\
DP Auditing  &  \cite{nasr2023tight},\cite{steinke2023privacy},\cite{jagielski2020auditing}, \cite{nasr2021adversary}, \cite{lu2022general}       &  \cref{sec:theory:aduiting}       \\
Learning Process Modifications  & \cite{abadi2016deep}, \cite{zhou2020private},  \cite{lee2018concentrated},  \cite{xu2020adaptive}, \cite{dormann2021not},  \cite{wang2019efficient}, \cite{ding2022differentially}, \cite{xiao2022differentially}, \cite{subramani2021enabling}, \cite{li2021large}, \cite{bu2022scalable}, \cite{bu2022differentially}, \cite{kairouz2020fast}, \cite{xiang2021differentially}, \cite{li2022private} ,\cite{li2022differentially}, \cite{wei2022dpis}, \cite{xiao2025trustworthy}\cite{zhang2025disk}         &  \cref{sec:theory:variants}       \\ 
Teacher-Student models     & \cite{papernot2016semi},  \cite{uniyal2021dp}, \cite{papernot2018scalable},  \cite{zhu2020private}, \cite{sun2020differentially} &
  \cref{sec:public_data:teacher}          \\ 
\bottomrule
\end{tabular}%
}
\end{table*}

\begin{table*}[]
\centering
\caption{Performance comparison among recent works that claim they have SoTA results. Note that these works usually report results for different $\varepsilon$ and datasets. Here we attempt to summarize them for two datasets: CIFAR-10 and ImageNet, and similar levels of privacy setting in terms of $\varepsilon$ and $\delta$. We observe that modifying existing model architectures, hyper-parameters tuning with training tricks, and gradient clipping techniques can result in the highest performance in terms of accuracy. That said, other directions are still worth exploring to achieve better performance in terms of running time or memory usage.}
\label{tab:sota}
\resizebox{\textwidth}{!}{%
\begin{tabular}{|c|c|c|c|c|c|c|}
\hline
Category &
  Reference &
  Dataset &
  Pre-train dataset &
  $\varepsilon$ &
  $\sigma$ &
  DP test acc. \\ \hline
Better privacy accounting &
  Ye et al. \cite{ye2022differentially} &
  \multirow{15}{*}{CIFAR-10} &
  \multirow{7}{*}{No} &
  3 &
  \multirow{15}{*}{$10^-5$} &
  69.30\% \\ \cline{1-2} \cline{5-5} \cline{7-7} 
Hyper-parameter selection &
  Dormann et al. \cite{dormann2021not} &
   &
   &
  7.42 &
   &
  70.10\% \\ \cline{1-2} \cline{5-5} \cline{7-7} 
\multirow{2}{*}{Model architecture search} &
  Cheng et al. \cite{cheng2022dpnas} &
   &
   &
  3 &
   &
  68.33\% \\ \cline{2-2} \cline{5-5} \cline{7-7} 
 &
  Remerscheid et al. \cite{remerscheid2022smoothnets} &
   &
   &
  7 &
   &
  73.50\% \\ \cline{1-2} \cline{5-5} \cline{7-7} 
Activation function modification &
  Papernot et al. \cite{papernot2021tempered} &
   &
   &
  7.53 &
   &
  66.20\% \\ \cline{1-2} \cline{5-5} \cline{7-7} 
Existing model modification &
  Klause et al. \cite{klause2022differentially} &
   &
   &
  8 &
   &
  82.50\% \\ \cline{1-2} \cline{5-5} \cline{7-7} 
Data augmentation &
  De et al. \cite{de2022unlocking} &
   &
   &
  8 &
   &
  81.40\% \\ \cline{1-2} \cline{4-5} \cline{7-7} 
Feature selection &
  Tram\`er and Boneh~\cite{tramer2020differentially}  &
   &
  ImageNet (unlabeled) &
  2 &
   &
  92.70\% \\ \cline{1-2} \cline{4-5} \cline{7-7} 
\multirow{2}{*}{Gradient clipping} &
  Bu et al. \cite{bu2022scalable} &
   &
  ImageNet1k &
  1 &
   &
  96.70\% \\ \cline{2-2} \cline{4-5} \cline{7-7} 
 &
  Bu et al. \cite{bu2022automatic} &
   &
  ImageNet (unlabeled) &
  2 &
   &
  92.70\% \\ \cline{1-2} \cline{4-5} \cline{7-7} 
Teacher-Student model &
  Zhu et al. \cite{zhu2020private} &
   &
  Subset of CIFAR-10 (unlabeled) &
  2.92 &
   &
  70.80\% \\ \cline{1-2} \cline{4-5} \cline{7-7} 
\multirow{5}{*}{Fine-tune technique} &
  Tang et al. \cite{tang2023differentially} &
   &
  Random process data &
  1 &
   &
  72.32\% \\ \cline{2-2} \cline{4-5} \cline{7-7} 
 &
  Cattan et al. \cite{cattan2022fine} &
   &
  ImageNet &
  1 &
   &
  95.00\% \\ \cline{2-2} \cline{4-5} \cline{7-7} 
 &
  Luo et al. \cite{luo2021scalable} &
   &
  CIFAR-100 &
  1.5 &
   &
  81.57\% \\ \cline{2-2} \cline{4-5} \cline{7-7} 
 &
  Amid et al. \cite{amid2022public} &
   &
  Subset of CIFAR-10 &
  3.51 &
   &
  67.03\% \\ \cline{2-7} 
 &
  Mehta et al. \cite{mehta2022large} &
  \multirow{2}{*}{ImageNet} &
  JFT-300M &
  4 &
  \multirow{2}{*}{$8*10^-7$} &
  81.70\% \\ \cline{1-2} \cline{4-5} \cline{7-7} 
Hyper-parameters tuning &
  Sander et al. \cite{sander2022tan} &
   &
  No &
  8 &
   &
  39.20\% \\ \hline
\end{tabular}%

}
\end{table*}
\section{Background \& Related Work}

To ensure the paper is self-contained, we provide background on supervised learning, ML reproducibility, and DPML. Readers may also be interested in recent surveys on DPML~\cite{ha2019differential,blanco2022critical,el2022differential,ponomareva2023dp,pan2024differential,zheng2025benchmarking,demelius2025recent}. 

\subsection{Supervised Learning}

A model maps data points (e.g., images) into (predicted) labels. The model itself is represented by a parameter vector and it is trained by minimizing a loss function that measures the discrepancy between predicted and actual labels of a training dataset. This is done using an optimization algorithm such as Stochastic Gradient Descent (SGD)~\cite{gower2019sgd}.

\paragraphb{Stochastic Gradient Descent (SGD)}
SGD is an iterative optimization procedure that has been extensively studied and has many variants~\cite{haji2021comparison}. Its mini-batch version is most commonly used and the one we refer to (unless otherwise stated) as SGD. It is illustrated in~\cref{alg:sgd}. Informally, the algorithm computes the gradient of the loss with respect to the parameter vector and iteratively updates the current solution accordingly until convergence.

\subsection{Transfer Learning}
Transfer learning is the process of adapting learned knowledge on one task or domain to another task or domain. We refer the reader to Weiss et al.~\cite{weiss2016survey} for a comprehensive discussion of the topic. For us, it suffices to note that the typical workflow for transfer learning is to use a model (e.g., a neural network) that is pre-trained on some dataset, change its last few layers (e.g., add a classification head), and then run the training process on the new task to fine-tune the existing pre-trained layers alongside with the new layers.

\subsection{ML Reproducibilty}
There are notable recent concerns regarding reproducibility in machine learning~\cite{kapoor2023leakage,heil2021reproducibility,pineau2021improving,semmelrock2023reproducibility}. For instance, Kapoor et al. \cite{kapoor2023leakage} highlight that the reproducibility crisis in ML, particularly the problem of data leakage. Tatman et al. \cite{tatman2018practical} suggest a taxonomy for reproducibility with three levels. Raff et al.~\cite{raff2019step} focus on independent reproducibility, where they implement 255 papers and record the features of each paper. They find that papers with a greater empirical focus are more reproducible. Pineau et al.~\cite{pineau2021improving} report on the NeurIPS 2019 reproducibility program and discuss the various factors that can lead to unreliable or false results.  
Heil et al.~\cite{heil2021reproducibility} propose reproducibility standards for machine learning in the life sciences.

\subsection{Differential Privacy and Variants}

Differential privacy (DP)~\cite{dwork2014algorithmic} is defined as follows.
\begin{definition}
    A randomized algorithm \dpalg{} is said to satisfy $(\varepsilon, \delta)$-{\em differential privacy} if for any neighboring datasets $D_0$, $D_1$ and any output set $S \subseteq {\rm Range}(\dpalg)$, it holds that:
        \[
            \pr{\dpalg(D_0) \in S} \leq \exp(\varepsilon) \ \pr{\dpalg(D_1) \in S} + \delta \ ,
        \]
\end{definition}
\noindent where probabilities are taken over randomness in \dpalg{}, and $\varepsilon > 0$ is called the {\em privacy budget}. The smaller $\varepsilon$ is the more stringent the privacy guarantee. We must also ensure that $\delta$ is sufficiently small. In the definition, $D_0$ and $D_1$ are neighboring if one can be obtained from the other by adding exactly one example.

An important concept for DP algorithms is sensitivity. Informally, sensitivity measures the maximum change in the output caused by the inclusion/exclusion of exactly one example. The (global) {\em sensitivity} of a function $g$ can be denoted as $\Delta g$ and is defined as 
\[
    \Delta g = \max || g(D_0) - g(D_1) || ,
\]
\noindent where the maximum is taken over pairs of neighboring datasets $D_0, D_1$ and $||\cdot||$ denotes a norm like the $l_1$-norm or the $l_2$-norm.

\paragraphb{R{\'e}nyi DP (RDP)}
There are several variants (both generalization and relaxations) of differential privacy~\cite{asoodeh2021three,kasiviswanathan2014semantics,desfontaines2020sok} such as approximate DP~\cite{dwork2006our}, R{\'e}nyi DP (RDP)~\cite{mironov2017renyi}. Among those DP variants, R{\'e}nyi DP is the most widely used. It is based on R{\'e}nyi divergence~\cite{renyi1961measures} and is defined as follows (Mironov~\cite{mironov2017renyi}). 
\begin{definition}
    A randomized algorithm \dpalg{} is said to be $(\alpha, \varepsilon)$-RDP with order $\alpha \geq 1$ and $\varepsilon \geq 0$ if for any neighboring datasets $D_0$, $D_1$, it has $ {\rm Div}_{\alpha}(\dpalg(D_0)|| \dpalg(D_1)) \leq \varepsilon$ where
    \begin{align*}
     & {\rm Div}_{\alpha}(\dpalg(D_0)|| \dpalg(D_1))  := \\
     &\frac{1}{\alpha-1} \cdot  \log  \mathop{\mathbb{E}}\limits_{Y \gets \dpalg(D_0)}
     \left[ \left(\frac{\pr{\dpalg(D_0)=Y}}{\pr{\dpalg(D_1)=Y}}\right)^{\alpha-1}\right] 
    \end{align*}
\end{definition}

RDP has several properties that make it useful in the context of machine learning such as its relationship with the Gaussian mechanism~\cite{wang2019subsampled,mironov2019r} and its composition properties. 
We refer the reader to Mironov~\cite{mironov2017renyi} for comprehensive coverage. However, an important property we highlight here is that we can convert an RDP guarantee to a (classical) DP guarantee. 
\begin{lemma}
If a randomized algorithm \dpalg{} is said to be $(\alpha, \varepsilon)$-RDP, then it also satisfy ($\varepsilon+\log(1/\delta)/(\alpha-1), \delta$)-DP for all $\delta \in (0,1)$.
\end{lemma}

\begin{algorithm}
\caption{SGD}
\label{alg:sgd}
{\small
\begin{algorithmic}
\Require Training dataset $D$, loss function $\mathcal{L}(\theta)$. Parameters: learning rate $\eta_t$, mini-batch size $L$. $N = |D|$ is the number of training data points.
\State \textbf{Initialize} $\theta_0$ randomly
\For{$t \in [T]$}
 \State Take a random mini-batch $L_t$ with probability $L/N$. 

\State \texttt{Compute gradient:}
\State $\mathbf{g}_t\leftarrow\nabla_{\theta}\mathcal{L}(\theta_t,L_t) $
\State \texttt{Descent step:}
\State $\theta_{t+1} \leftarrow \theta_t-\eta_t{\mathbf{g}}_t$
\EndFor
\Ensure $\theta_T$ 

\end{algorithmic}
}
\end{algorithm}

\begin{algorithm}
\caption{DP-SGD (Abadi et al.~\cite{abadi2016deep})}
\label{alg:dpsgd}
{\small
\begin{algorithmic}
\Require Training data ${x_1,...,x_N }$, loss function $\mathcal{L}(\theta)=\frac{1}{N}\sum_i\mathcal{L}(\theta,x_i)$. Parameters: learning rate $\eta_t$, noise scale $\sigma$, mini-batch/lot size $L$, gradient norm bound $C$.
\State \textbf{Initialize} $\theta_0$ randomly
\For{$t \in [T]$}
\State Take a random sample $L_t$ with probability $L/N$
\State \texttt{Compute gradient:}
\State For each $i\in L_t$, compute $\mathbf{g}_t(x_i)\leftarrow\nabla_{\theta}\mathcal{L}(\theta_t,x_i) $
\State \texttt{Clip gradient:}
\State $\bar{\mathbf{g}}_t(x_i) \leftarrow \mathbf{g}_t(x_i)/\mathrm{max}(1,\frac{||\mathbf{g_t(x_i)}||_2}{C})$
\State \texttt{Add noise:}
\State $\tilde{\mathbf{g}}_t \leftarrow \frac{1}{L} \sum_i(\bar{\mathbf{g}}_t(x_i)+\mathcal{N}(0,\sigma^2C^2\mathbf{I}))$
\State \texttt{Descent step:}
\State $\theta_{t+1} \leftarrow \theta_t-\eta_t\tilde{\mathbf{g}}_t$
\EndFor
\Ensure $\theta_T$ and  and compute the overall privacy cost $(\varepsilon, \delta)$ using a privacy accounting method
\end{algorithmic}
} 
\end{algorithm}

\subsection{Differentially Private Machine Learning}
There are numerous randomized learning algorithms that meet differential privacy such as output perturbation~\cite{dwork2006our} or objective perturbation~\cite{chaudhuri2011differentially}. However, the most widely used technique is the Differentially Private Stochastic Gradient Descent (\dpsgd) of Abadi et al.~\cite{abadi2016deep}.

\paragraphb{\dpsgd}
In essence, \dpsgd{} computes gradients and applies gradient updates in a similar fashion as \sgd{}, but uses the Gaussian mechanism to add noise to the aggregate mini-batch gradient. However, since the sensitivity of the mini-batch gradient to a single example is unbounded, the per-example gradients are clipped prior to aggregation to ensure that their $l_2$-norm is smaller than a clipping threshold $C>0$. This guarantees that the sensitivity of the mini-batch gradient sum is at most $C$. In other words, the noisy gradient is calculated as:
\[
    \bar{g} = \frac{1}{L} \sum_i  {\rm clip}_C(g_i) + \mathcal{N}(0, \sigma^2 C^2 \mathbf{I}) \ ,
\]
where $L$ is the number of examples in the mini-batch, $g_i$ is the gradient vector of example $i$, and $\sigma$ is the noise level. For completeness, we provide a complete description of the \dpsgd{} algorithm in~\cref{alg:dpsgd}. Since it is an iterative algorithm where gradients are computed multiple times over the same data, careful privacy budget account and composition is required~\cite{abadi2016deep}. 
A notable drawback of per-example gradient clipping as used in DP-SGD is processing speed. The running time for one epoch with DP-SGD is between 10 to 30 times slower than with SGD ~\cite{bao2022importance,subramani2021enabling}.

\paragraphb{Privacy-utility trade-off}
There is a natural tradeoff between privacy and utility when training the model with \dpsgd{}. This is because the smaller the privacy budget $\varepsilon$ the more noise needs to be added to the clipped gradients and thus the more distortion the training process will experience.
In this paper, we think of privacy as the total privacy budget $\varepsilon$ and of utility as a combination of test accuracy of the trained model and also training time and memory usage.

\section{Architectures, Features, Hyperparameters and Augmentations}\label{sec:tuning}
Best practices for machine learning in the non-private setting do not typically yield the best results for DPML~\cite{abadi2016deep,papernot2020making,papernot2021tempered,tramer2020differentially,bao2022importance,singhal2021privately,priyanshu2021efficient,cheng2022dpnas,remerscheid2022smoothnets}. This is why significant research has gone into tuning model architectures, hyperparameter optimization and feature selection for DPML.

\subsection{Tuning Model Architectures}\label{sec:experiment_direction:model}
Several researchers have pointed out that model architectures that perform well in the non-DP setting may have poor performance in the DP setting~\cite{papernot2020making,morsbach2021architecture,kurakin2022toward,bao2022importance}. Therefore, carefully selecting model architectures is necessary to achieve the best results.

\paragraphb{Architecture search}

Priyanshu et al.~\cite{priyanshu2021efficient} evaluate Evolutionary, Bayesian, and Reinforcement learning search algorithms, finding them superior to grid search. Bao et al.~\cite{bao2022importance} use a genetic algorithm for model architecture search, mutating model elements as genes and evaluating with the Laplace mechanism for privacy. Cheng et al.~\cite{cheng2022dpnas} focus on a narrower search space, introducing a changeable NAS Cell in a CNN model. The authors optimize this cell using Reinforcement Learning. Remerscheid et al.~\cite{remerscheid2022smoothnets} empirically select model components to design SmoothNet, which achieves better DP performance than other well-known models.

\paragraphb{Activation function}
%
Papernot et al.~\cite{papernot2021tempered} observe that DP-SGD causes model activations to explode, leading to information loss during gradient clipping. The authors propose using a bounded activation function, a tempered sigmoid, to replace ReLU. Experiments show improved privacy-utility trade-offs with this activation function.

\paragraphb{Loss function}
%
Wang et al.~\cite{wang2021dplis} propose smoothing the loss function using randomized smoothing, improving performance in image and language tasks. Shamsabadi et al.\cite{shamsabadi2021losing} present a DP-SGD tailored loss function combining per-example sum squared error, a focal loss variant, and a regularization penalty. This approach is also thought to smooth out the loss surface, enhancing DP noise tolerance.

\paragraphb{Tuning existing architectures}
Instead of searching for architectures from scratch, one idea is to take a well-established architecture and make it DP-friendly. For example, Residual networks (ResNet~\cite{he2016deep}) are very popular in computer vision tasks and won several victories in image challenges such as ImageNet (ILSVR) challenge. 

Klause et al.~\cite{klause2022differentially} proposed an architecture modification on ResNet, adding a normalization layer after the residual block called ScaleNorm. The authors claim that ScaleNorm can improve the speed of convergence, and achieve SOTA performance on CIFAR-10 when trained from scratch. Sander et al.~\cite{sander2022tan} propose changing the order of layers in a ResNet block. They show experimentally that changing the order of activation and normalization layers has a significant impact on performance. Specifically, that using normalization before ReLU leads to improved performance compared to using ReLU before batch normalization.

\subsection{Feature Selection}\label{sec:experiment_direction:train}
%
Reducing data dimensionality can mitigate the curse of dimensionality and enhance performance. DP-PCA was introduced by Dwork et al.~\cite{dwork2014analyze}. Abadi et al.~\cite{abadi2016deep} in the original \dpsgd{} paper, demonstrates that PCA can boost performance by reducing model parameters.

Aside from PCA, Tram\`er and Boneh~\cite{tramer2020differentially} propose to use handcrafted features and show that a model trained on these features can outperform models trained from scratch without such features. Specifically, they use a Scattering Network as a feature extractor with Group Normalization. They combine it with the linear model and deep models in experiments and show that all of them outperform models that do not use feature selection. We evaluate their methods in our reproducibility experiments (\cref{sec:exp:train_from_scratch}).

Bao et al.~\cite{bao2022importance} describe three feature selection methods depending on the available privacy budget, pointing out that using all features does not always yield the best model. Singhal et al.\cite{singhal2021privately} propose a dimensionality reduction method that projects data into low-dimension linear subspace.

\subsection{Hyperparameter Tuning}\label{sec:hyperparam}

Researchers have pointed out the importance of tuning hyperparameters (e.g., learning rate, regularization constant, batch size, etc.) in the DP setting~\cite{bao2022importance,kurakin2022toward,de2022unlocking,sander2022tan}. 

Kurakin et al.~\cite{kurakin2022toward} find that hyperparameters (i.e, suitable learning rate, larger batch size, or epochs) have a significant impact on DP performance.

Dormann et al.~\cite{dormann2021not} investigate that the inherent sampling noise in SGD and Gaussian noise in DP-SGD is equivalent to achieving privacy. They propose a novel method for tuning hyperparameters that uses a larger batch size and high noise multiplier to achieve SoTA performance. We include their method in our reproducibility experiments (\cref{sec:exp:train_from_scratch}).

\paragraphb{Costs of hyperparameter tuning}
Although hyperparameter tuning is crucial, it can be computationally expensive, and ideally, its privacy effect should be accounted for.

Sander et al.~\cite{sander2022tan} proposed a method for reducing the computational cost of hyperparameter tuning with DP-SGD. The method involves using the Total Amount of Noise (TAN) and a scaling law. TAN is defined as the sum of the variances of the noise added to each gradient update during training and represented as follows:
\[
\eta ^2 = \frac{1}{\Sigma^2} = \frac{q^2 S}{2 \sigma ^2} \ ,
\]
where $\eta$ is the signal-to-noise ratio, $\Sigma$ is the TAN, $q$ is simple rate, $S$ is the step size and $\sigma$ is noise level.

The authors argue that $\varepsilon$ is directly related to TAN and can be minimized, which means that TAN can be used to predict privacy guarantees. Furthermore, the scaling law can be leveraged to estimate the performance of large batch sizes using smaller batch sizes. In experiments, Sander et al.~\cite{sander2022tan} show that their method can find optimal hyperparameters, leading to new SoTA performance on ImageNet.

Ideally, the privacy cost of hyperparameter tuning should also be considered. Papernot and Steinke~\cite{papernot2021hyperparameter} highlight that hyperparameter settings can leak private information and offer theoretical tools to ensure privacy during tuning. Wang et al.~\cite{wang2023dp} introduce DP-HyPO, an adaptive hyperparameter optimization framework that combines adaptive search with DP and can refine sampling distributions adaptively based on previous runs while maintaining strict DP guarantees.

\subsection{Data Augmentation}

\label{sec:misc:aug}

De et al.~\cite{de2022unlocking} advocate for the use of large batches and replace batch normalization with group normalization, and weight standardization. They also propose to create several {\em self-augmentations} of each example and then average their gradients before clipping. This does {\em not} affect the privacy analysis because it happens prior to clipping (thus any given example only affects one clipped gradient value per mini-batch). This is particularly effective and results in a new SoTA on CIFAR-10 when trained from scratch.

Building on this, Bao et al. \cite{bao2024dp} introduced \selfmix{} a technique that employs Mixup~\cite{zhang2018mixup} for self-augmentation of individual training samples, resulting in state-of-the-art (SoTA) performance in both training models from scratch and fine-tuning pre-existing models. A second method, called \diffmix, aims to further enhance performance by using a text-to-image diffusion model to create class-specific synthetic examples. These synthetic samples are then mixed up with actual training data, achieving new SoTA performance levels without incurring additional privacy costs. These two techniques have established new SoTA benchmarks across various image datasets. 

Park et al.~\cite{park2024distribution} propose synthesizing a large, diverse dataset from a small set of in-distribution public data using diffusion models to improve warm-up training and boost the final utility of privately trained models.
\begin{table}[h]
\caption{Different types of gradient clipping techniques}
\label{tab:clipping}
\resizebox{\columnwidth}{!}{%
\begin{tabular}{ll}
\toprule
Type of clipping techniques    & Reference                                                                                           \\ \midrule
Improvements on basic clipping &~\cite{chen2020understanding},~\cite{lee2021scaling},~\cite{bu2021fast},\cite{bu2022scalable}      \\ 
Adapt clipping &
 ~\cite{yu2018improve},~\cite{pichapati2019adaclip},~\cite{andrew2021differentially},~\cite{du2021dynamic},~\cite{lin2022understanding},~\cite{golatkar2022mixed},~\cite{xia2023differentially} \\ 
Auto clipping                  &~\cite{bu2022automatic} ,~\cite{yang2022normalized}                                                  \\ 
Layer(Group) clipping          &~\cite{abadi2016deep},~\cite{mcmahan2018learning},~\cite{van2018three},~\cite{liu2021differentially} \\ 
Global clipping                &~\cite{bu2021convergence}                                                                            \\ \bottomrule
\end{tabular}%
}
\end{table}
%

\section{All About Clipping} \label{sec:clipping}
Gradient clipping is an essential part of \dpsgd{} and also the focus of numerous techniques claimed to boost DP model performance. These can be categorized into five classes and we summarize them in~\cref{tab:clipping}. In this section, we provide a brief overview of existing clipping techniques.

\subsection{Basic Clipping \& Limitations}
The original gradient clipping process proposed by Abadi et al.~\cite{abadi2016deep} is defined in~\cref{eq:basic_clip}. We call it {\em basic clipping}. 
\begin{equation}\label{eq:basic_clip}
\bar{\mathbf{g}}_t \leftarrow \mathbf{g}_t \cdot \mathrm{min}(1,\frac{C}{||\mathbf{g_t}||_2}) \ ,
\end{equation}
where $\mathbf{g}_t $ is the original gradient vector at iteration $t$, $\bar{\mathbf{g}}_t$ is the clipped gradient vector, and $C$ is the clipping threshold. 

In a nutshell, clipping preserves the gradient vector whenever its $l_2$-norm is bounded by the clipping threshold $C$. When it is not, the gradient vector is first normalized and then re-scaled by a factor of $C$, preserving the direction but reducing magnitude.

Instead of considering $C$ as an independent parameter, De et al. \cite{de2022unlocking} merge the clipping factor into the learning rate and tune the two parameters together to simplify tuning without altering the privacy guarantee.

\paragraphb{Limitations of basic clipping}
Basic clipping in \dpsgd{} can distort gradient vectors, impacting convergence. Chen et al.~\cite{chen2020understanding} found that clipping bias affects convergence and propose adding Gaussian noise before clipping to counteract this. Koloskova et al. \cite{koloskova2023revisiting} detail how different clipping thresholds affect convergence and aim for improved convergence guarantees through adaptable thresholds. Meanwhile, Xiao et al. \cite{xiao2023geometry} study optimal Gaussian noise with hybrid clipping, introducing a "twice sampling" method for better utility-privacy trade-offs, and exploring optimal noise for RDP in high dimensions.

\paragraphb{Beyond basic clipping}
A consequence of basic clipping is that per-example gradient clipping slows down training and can take up significant memory. Lee et al.~\cite{lee2021scaling} propose a modification on basic clipping that uses the ``auto-differentiation'' library in deep learning frameworks to compute the norms of per-example gradients. This method can be implemented on different types of neural networks and show significant speed-ups. 

Bu et al.~\cite{bu2021fast} propose a fast gradient clipping that approximates per-sample gradient norms instead of computing them by using Johnson–Lindenstrauss projections which could save significant $30 \times$ time and use a similar memory footprint as non-private optimization.

\subsection{Adaptive Clipping}
Instead of using a fixed clipping threshold like basic clipping, Adapt clipping~\cite{yu2018improve, pichapati2019adaclip, andrew2021differentially, du2021dynamic, lin2022understanding, golatkar2022mixed} seeks to use different clipping thresholds for different iterations to improve model's performance. 

Yu et al.~\cite{yu2018improve} note gradients tend to zero in convergent algorithms, so they linearly decrease the clipping threshold to half its original value. Pichapati et al.~\cite{pichapati2019adaclip} introduce AdaClip, a coordinate-wise method that requires less DP noise for a given privacy budget. Andrew et al.~\cite{andrew2021differentially} adapt the clipping threshold using gradient quantile approximations. Du et al.~\cite{du2021dynamic} propose a dynamic threshold that decreases at every iteration, adjusting noise scaling accordingly. Lin et al.~\cite{lin2022understanding} suggest a method to find the optimal threshold on public data with a greedy algorithm, and a decay gradient clipping threshold using a decay function. Golatkar et al.~\cite{golatkar2022mixed} employ public data to gauge the 90th percentile of gradient norms, using this insight for adaptive clipping on private data. Xia et al.~\cite{xia2023differentially} introduce DP-PSAC, which uses an adaptive weight function instead of a constant clipping norm, narrowing the deviation between true and batch-averaged gradients.

\subsection{Efficient Clipping}

\paragraphb{Auto clipping}\label{sec:clipping:auto}
No matter the fixed clipping threshold or adaptive clipping threshold, there are always one or more hyper-parameters that need to tune. Bu et al.~\cite{bu2022automatic} and Yang et al.~\cite{yang2022normalized} proposed a new clipping technique called Auto clipping which can be defined as follows:
\[
\bar{\mathbf{g}}_t \leftarrow \mathbf{g}_t \cdot \frac{1}{||\mathbf{g_t}||_2 + \gamma} \ .
\]

This method removes the clipping threshold instead of using methods to find it. The idea is to incorporate a parameter $\gamma$ which can be relatively stable for different sets of hyperparameters, to preserve the gradients' magnitude information. The authors show that the convergence of DP-SGD with Auto clipping is the same as the standard SGD. In experiments, DP-SGD with Auto clipping achieves slightly better performance than DP-SGD with basic clipping. 

\paragraphb{Mixed ghost clipping}\label{sec:clipping:Mixed}
By combining and extending previous works~\cite{goodfellow2015efficient,lee2021scaling,li2021large}, Bu et al.~\cite{bu2022scalable} propose a technique called mixed ghost gradient clipping which does not require computing per-sample gradients.  
Their method yields computation time and memory close to non-private optimization. They claim to achieve a new SOTA performance on CIFAR-10 and CIFAR-100.

\subsection{Group Clipping}
This subsection surveys group-based gradient-clipping strategies—layer-wise, global, and batch-level—that refine standard per-sample clipping by assigning tailored thresholds to structural or statistical subsets of the gradients, thereby reducing utility loss while preserving differential-privacy guarantees.

\paragraphb{Layer clipping}
Layer clipping~\cite{abadi2016deep,mcmahan2018learning,van2018three} sets unique clipping thresholds for different model layers, recognizing that each layer has distinct gradient norms. In contrast, Liu et al.~\cite{liu2021differentially} introduce a method dividing gradients into $k$ groups, each with its own clipping threshold, showing reduced gradient loss compared to basic clipping. Xu et al.~\cite{xu2020removing} present DPSGD-F, a group clipping technique with an adaptive threshold, adjusting each group's sample contribution to equalize utility loss across groups. Their experiments demonstrate improved DP performance.

\paragraphb{Global clipping}
Bu et al.~\cite{bu2021convergence} present global clipping techniques that assign either 0 or 1 as the clipping factor to all gradients. This method keeps small gradients and removes large ones, typically from noisy samples, aiming to maintain gradient direction and minimize bias.

\paragraphb{Batch clipping}
Xiao et al. \cite{xiao2023theory} analyze gradient clipping theoretically, suggesting its bias is often underestimated due to the gradient's sampling noise. Based on this, they propose techniques, including inner-outer momentum and Batch Clipping, using public data to inform batch sample clipping. They also offer other strategies to boost DPML performance.

\section{Transfer Learning \& Fine-tuning}\label{sec:public_data}

A model pre-trained on a public dataset can be used as a starting point for DP training. A typical methodology is to reuse the weights of the pre-trained model and then fine-tune a new model on the sensitive dataset with \dpsgd{}. 

The benefit comes from the use of public datasets for which the privacy cost is {\em not} taken into account. The assumption is that the pre-training data is public or otherwise not sensitive. This assumption becomes increasingly dubious the more similar the pre-training and fine-tuning datasets are. At the same time, the more similar the pre-training and fine-tuning datasets are the better the expected performance.

\subsection{Transfer Learning, Domains, and Tasks}\label{sec:public_data:fine-tune}

It is unclear whether pre-training the model on public data is still beneficial when there is a significant domain gap between private data and public data for image tasks. However, many papers~\cite{abadi2016deep,amid2022public,golatkar2022mixed} that focus on image tasks tend to assume that public data and private data have similar distributions.

By contrast, for language tasks, a large model pre-trained with generic public data appears to still yield good performance when fine-tuned with out-distribution private data~\cite{li2021large,yu2021differentially,shi2022just,hoory2021learning,kerrigan2020differentially}.   

However, using public data in DPML also brings some concerns. Tram\`er et al. \cite{tramer2022considerations} discuss improving DPML through transfer learning from extensive non-private datasets. They express concerns over differential privacy in web-acquired data and the challenges of outsourced data in large pre-trained models, urging a closer look at privacy implications in public pre-training.

\subsection{Parameter-Efficient Fine-Tuning Techniques}
Mehta et al.~\cite{mehta2022large} find that increased pre-training data and bigger models enhance performance. They also observe improvements using large batch sizes, LAMB optimization~\cite{you2019large}, and initializing the final layer's weights to zero for fine-tuning with DP.

Luo et al.~\cite{luo2021scalable} propose a new fine-tuning technique. Their techniques fine-tune the normalization layer and percentage of parameters in convolution layers that have high magnitude. This results in SoTA performance. 

Cattan et al.~\cite{cattan2022fine} propose fine-tuning the first and last layers to improve performance (an increase of $3.2\%$ accuracy compared to fine-tuning the whole model). They claim to achieve SoTA performance.

Bu et al.~\cite{bu2022differentiallyb} introduce DP-BiTFiT which fine-tunes only the model's bias using DP-SGD. With only $0.1\%$ parameter changes, this technique achieves leading results. The authors propose a two-phase training for image tasks, first full-model fine-tuning and then DP-BiTFiT.
Li et al.~\cite{li2021large} describe ghost clipping for fine-tuning, noting enhanced results with larger models and optimal hyperparameters.
Yu et al.~\cite{yu2021differentially} present a meta-framework, suggesting techniques like low-rank adaptation can boost fine-tuning in large models.
Lowy et al. \cite{lowy2023optimal} explore optimizing DP learning with public data, highlighting its benefits in reducing sample complexity.

For language tasks, 
Hoory et al.~\cite{hoory2021learning} introduce the DP Word Piece algorithm for language tasks, adding noise to word histograms and leveraging parallel training on TPUs. Yu et al.~\cite{yu2023selective} champion ``selective pre-training'' on public data, later finetuning on private datasets, benefiting both privacy and model efficiency. Mireshghallah et al.~\cite{mireshghallah2022differentially} focus on compressing models like BERT to 50\% sparsity using DPKD and DPIMP, achieving competitive performance on GLUE.

\subsection{DP-Aware Pre-Training and Representation Learning}

This subsection focuses on works that first shape the feature extractor (synthetic images, captioning, robustness objectives) so that subsequent DP fine-tuning is both easier and more accurate.

Tang et al. \cite{tang2023differentially} introduce a method to pre-train models on random process data~\cite{baradad2021learning,baradad2022procedural}, eliminating reliance on existing public datasets. Their three-phase process starts with pre-training the model's feature extractor on this data. Next, they train the classification layer solely on private data, freezing the feature extractor. In the last phase, the entire model is fine-tuned on private data. Their approach achieves top-tier results, underlining its effectiveness. 

Similar to Tang et al. \cite{tang2023differentially}, Yu et al. \cite{yu2023vip} propose a method to train foundation vision models with differential privacy using masked autoencoders and DP-SGD. They pretrain MAE on synthetic images which is random process data as \cite{tang2023differentially} and adapt MAEs for DP by utilizing an instance-separable loss function that aligns well with the per-sample gradient computations required for DP-SGD. This method enables the effective training of privacy-preserving models on large-scale, uncurated datasets, maintaining competitive performance on standard vision tasks.
Although sharing a similar idea with Tang et al. \cite{tang2023differentially}, Yu et al. \cite{yu2023vip} focuses on a different training situation: private pre-training. 

Sander et al. \cite{sander2024differentially} introduces a representation learning method using image captioning, which enhances model performance by utilizing text captions for better supervision and information extraction under privacy constraints. The authors develop a DP-Cap model that significantly outperforms previous models in learning high-quality image features from a large-scale dataset (LAION-2B). 

Wang et al. \mbox{\cite{wang2024dpadapter}} propose DPAdapter, a pre-training technique that enhances a model's parameter robustness against DP noise by improving sharpness-aware minimization (SAM) \mbox{\cite{foret2020sharpness}} with a two-batch strategy for more accurate perturbation estimation, making the model more resilient during downstream DP fine-tuning.

Besides getting a better starting point, public data with a similar distribution as private data can also give some insight during fine-tuning with DP-SGD.
Golatkar et al.~\cite{golatkar2022mixed} use an adaptive method that processes gradients in a low-dimensional space, enhancing DP fine-tuning. Amid et al.~\cite{amid2022public} generate loss based on public data and compare it to the loss on private data. They let the fine-tuned model take smaller gradient steps with less noise when the public loss grows quickly.

\section{Other Techniques}\label{sec:theory}
This section surveys other techniques to improve privacy-utility from different perspectives.

\subsection{Sparse Training} \label{sec:misc:sparse}
Some research propose to mitigate the impact of clipping and noise on the gradient by reduce the effective size of gradient vectors. We categorize these methods into five classes summarized in~\cref{tab:gredient_size}.
\begin{table}[t]
\centering
\caption{Different types of gradient size reduction techniques}
\label{tab:gredient_size}
\resizebox{0.8\columnwidth}{!}{%
\begin{tabular}{ll}
\toprule
Type of gradient size reduction & Reference                                        \\ \midrule
Random projection             &~\cite{zhu2021differentially},~\cite{kasiviswanathan2021sgd},~\cite{sidahmed2021efficient},~\cite{bao2022importance} \\
Low-rank matrix approximation &~\cite{yu2021large},~\cite{ito2022scaling},~\cite{yu2021large}                                                       \\
Private selection gradient      &~\cite{zhang2021wide}                             \\
Subspace projection             &~\cite{asi2021private},~\cite{nasr2020improving},~\cite{feng2023spectral} \\
Public data utilization         &~\cite{zhou2020bypassing},~\cite{yu2020not},~\cite{gu2023choosing}        \\ \bottomrule
\end{tabular}%
}
\end{table}

\paragraphb{Random projection}
Random projection of the gradient has been proposed in traditional (non-DP) model training~\cite{nedic2010random,lee2013distributed}. 
Zhu and Blaschko~\cite{zhu2021differentially} introduce a DP training version that employs a random freeze mask, updating only a subset of gradients each iteration. This reduces noise addition, leading to enhanced DP performance, with added memory and communication efficiency benefits.
Kasiviswanathan~\cite{kasiviswanathan2021sgd} takes a similar approach but uses a subgaussian or sparse Johnson–Lindenstrauss matrix for random gradient projection instead of a freeze mask.

Sidahmed et al.~\cite{sidahmed2021efficient} introduce "partial training," freezing parts of the neural network with the most parameters. Based on performance, they adjust which parts are frozen. Their method shows better results with smaller privacy budgets than full-network training. Similarly, Bao et al.~\cite{bao2022importance} suggest ``random weights training,'' initializing and freezing the initial layers' weights.

\paragraphb{Low-rank matrix approximation}
Using low-rank matrices to represent a high-rank gradient matrix is a popular gradient size-decreasing method~\cite{pitaval2015convergence,ye2021global}. To implement it in DP training, Yu et al.~\cite{yu2021large} use two low-rank matrices to represent the gradient. By only adding noise to low-rank matrices, they can get better DP performance. Finally, they will reconstruct the gradient using noise low-rank matrices. Based on that work, Ito et al.~\cite{ito2022scaling} also generate two low-rank matrices but before adding DP noise, they will set the unimportant gradient of low-rank matrices to 0. So compared to Yu et al.~\cite{yu2021large}, they further decrease gradient size. These works show improved DP performance when the privacy budget is limited.

\paragraphb{Private selection gradient}
Current neural networks are often over-parameterized and there exists some sparse in the model that can be compressed. Based on this idea, it is natural to consider only updating the important subset of the gradient. To make the selection process differential private, private selection techniques need to be implemented. Zhang et al.~\cite{zhang2021wide} firstly clip the gradient and then use a private selection technique to generate a gradient mask that only keeps important gradients. Then, they do a second gradient clipping and add DP noise to it. 

\paragraphb{Subspace projection}
Subspace projection methods focus on projecting gradients to subspace which can still preserve its important information. Asi et al.~\cite{asi2021private} introduce a private version of AdaGrad with non-isotropic clipping and noise, projecting gradients into an ellipsoid space. Nasr et al.~\cite{nasr2020improving} encode gradients into a smaller vector space using different noise distributions for improved utility and employ a denoising mechanism to scale noisy gradients, achieving comparable results to DP-SGD on MNIST. Feng et al. \cite{feng2023spectral} present Spectral-DP, projecting DP noise in the spectral domain with spectral filtering, aiming to reduce noise magnitude.

\paragraphb{Public data}
Using public data information can assist in gradient reduction. Zhou et al.~\cite{zhou2020bypassing} compute the top-k eigenspace from public data for gradient projection, updating weights with projected noisy gradients. Yu et al.~\cite{yu2020not} employ unlabeled public data to form an anchor subspace, then project and perturb gradients within this subspace for differential privacy.

The choice of appropriate public data is crucial. Gu et al. \cite{gu2023choosing} present an algorithm to select suitable public datasets for private ML tasks. By leveraging the Gradient Subspace Distance (GSD) to measure dataset differences, the method aims to reduce noise. GSD's effectiveness is evaluated across various models, and found to be enhancing the private training of large-scale models.

\subsection{Better Privacy Accounting}\label{sec:theory:account}
%
There is considerable literature on proposing tighter privacy bounds through composition. This allows the learning algorithm to perform additional iterations while using the same privacy budget and noise level. Asoodeh et al.~\cite{asoodeh2020better} derive a tighter privacy bound based on a joint range of two $f$-divergences. $f$-divergence~\cite{harremoes2011pairs} is a function to measure the difference between two probability distributions. By using this method, the authors optimally convert from RDP to DP which can be applied to the moments accountant, allowing for 100 additional training iterations for the same privacy parameters. 

A series of recent studies use an accounting method based on the Fast Fourier Transform (FFT) method~\cite{cooley1965algorithm}. The FFT-based account method was first proposed by Koskela et al.~\cite{koskela2020computing}. The proposed accountant method is based on an integral formula numerical approximation which is realized by discretizing the integral and assessing discrete convolutions with the FFT algorithm. They also show in the experiments that their method has better performance than moments accountant and other accountant methods in terms of bound tightness and running time.
Koskela et al.~\cite{koskela2021tight} propose an accounting method that is based on the privacy loss distribution formalism by using the FFT-based account method. They conduct an error analysis to obtain a tighter bound and they show that their approach can decrease $75\%$ noise variance.
Koskela et al.~\cite{koskela2021computing} extend the FFT-based accounting method to heterogeneous composition to compute tight privacy bound and conduct an error analysis to produce a better complexity bound. Their experiments show that their method can allow $1.5 \times$ more compositions than the original moments accountant.
Gopi et al.~\cite{gopi2021numerical} propose a new algorithm based on previous work~\cite{koskela2021tight} that can compute the privacy composition curve more efficiently. Using the coupling approximation, their proposed algorithm shifts the discretized random variables and enhances the approximation procedure. They also improve the truncation approach by offering a tighter tail bound of the privacy loss random variables. Their new algorithm is fast at least $k$ times than previous work~\cite{koskela2021tight} ($k$ is the number of compositions).

A different way to achieve tighter privacy bounds is to reconsider traditional assumptions of differential privacy machine learning. In the original \dpsgd{} setting, all iterations' gradients are implicitly assumed to be visible to the adversary. Thus privacy accounting protects the entire training process's internal state, and not just the final model or iteration. 

However, this assumption may not be realistic in practice as model owners may train their models on isolated systems and only publish the model after the end of the training process. To capture this, several works propose the idea of {\em hidden-state analysis} where only the last iteration of training is observed by the adversary.

Feldman et al.~\cite{feldman2018privacy} demonstrate privacy guarantees can be amplified by not releasing the intermediate results. They also show that this new analysis can be applied to various applications such as distributed SGD, Multi-query settings, and Public/private data settings. 

Chourasia et al.~\cite{chourasia2021differential} propose a framework to analyze the dynamics privacy loss of noisy gradient descent algorithms based on a pair of continuous-time Langevin diffusion processes. They provide a tighter bound on the R\'enyi divergence than composition bounds under the assumption that only the last iteration information is released and the loss function is smooth and strongly convex. 

Extending previous work from gradient descent to SGD, Ye and Shokri~\cite{ye2022differentially} provide a tighter bound relying on the post-processing phenomenon for noisy mini-batch SGD. Their coverage bound can be implemented on convex loss function and is suitable for different batch sampling methods so it can also be implemented with \dpsgd{}. 

\subsection{DP Auditing} \label{sec:theory:aduiting}

Jagielski et al.~\cite{jagielski2020auditing} introduce a novel data poisoning attack to evaluate DP-SGD in real-world settings, comparing practical privacy to the theoretical estimates. 
Nasr et al.~\cite{nasr2021adversary} propose an approach to assess the actual privacy levels of DP-SGD using a conceptual adversary. They aim to set lower bounds on the likelihood of succeeding in a distinguishing game, a metric for data privacy. Their results indicate that in real-world scenarios, the actual privacy provided may exceed theoretical guarantees.

Lu et al.~\cite{lu2022general} present ML-Audit, a framework designed to statistically audit the privacy assurances of DPML. 
Nasr et al. \cite{nasr2023tight} introduce an enhanced auditing scheme for assessing the privacy of machine learning algorithms under differential privacy. Their approach requires just two runs and offers precise privacy estimates for non-adversarially curated datasets.
Steinke et al.~\cite{steinke2023privacy} reduce the computational overhead linked to privacy auditing. They show how to estimate a lower bound of $\varepsilon$ in a single run.

Hu et al. \cite{hu2025empirical} introduce the concept of ``empirical privacy variance,'' showing that models with the same theoretical DP guarantee can have significantly different practical privacy leakage based on hyperparameter choices.

There is also a line of research that focuses on auditing the privacy of Large Language Models (LLMs)~\cite{panda2025privacy, marek2024auditing, chard2024auditing,rathod2025privacy,yao2024survey,li2024llm,duan2024membership,meeus2025sok}.  
For example, Panda et al.~\cite{panda2025privacy} insert rare ``canary'' tokens to amplify memorization, enabling simple black-box membership-inference audits that reveal leakage even in DP-trained LLMs. These tests are lightweight and model-agnostic, and thus provide a way for practitioners to catch privacy bugs and estimate real-world risk.

\subsection{Learning Algorithm Modifications} \label{sec:theory:variants}
Researchers have proposed variants of \dpsgd{} to overcome some of its limitations such as slow convergence and long running time and (or) to achieve superior privacy and utility trade-offs. We summarize these variants in~\cref{tab:variants}.
\begin{table}[t]
\centering
\caption{Variants of DP-SGD}
\label{tab:variants}
\resizebox{0.75\columnwidth}{!}{%
\begin{tabular}{ll}
\hline
Type                                & Reference                                                \\ \hline
Adaptive optimization                 &~\cite{zhou2020private}                                   \\ 
Adaptive noising   &~\cite{lee2018concentrated},\cite{xu2020adaptive},\cite{dormann2021not},~\cite{xiang2021differentially}       \\ 
Improved accounting &~\cite{wang2019differentially},\cite{ding2022differentially}                  \\ 
Variants of DP-SGD &~\cite{xiao2022differentially,xiao2025trustworthy,wei2022dpis}                            \\
DP-SGD acceleration                   &~\cite{subramani2021enabling},\cite{bu2022differentially} \\ 
Public data utilization          &~\cite{kairouz2020fast},~\cite{li2022private} \\ \hline
\end{tabular}%
}
\end{table}

\paragraphb{Adaptive optimization} 
Shortening the number of training iterations directly reduces the overall privacy budget consumed. However, the rate of learning has to increase to compensate for having fewer learning iterations. Adaptive optimization techniques such as AdaGrad~\cite{duchi2011adaptive}, Adam~\cite{kingma2014adam}, and RMSProp~\cite{tieleman2012lecture} change the learning rate adaptively based on second-moment gradient information. These techniques have better performance than SGD in the non-DP setting. Therefore, when switching to the DP setting, it is natural to consider replacing DP-SGD with differential privacy adaptive optimization. Zhou et al.~\cite{zhou2020private} propose DP Adam and DP RMSProp and prove their DP guarantee. They also represent their convergence rate of them and evaluate their performance. They use experiments to show that DP adaptive optimization outperforms DP-SGD.   

Li et al. \cite{li2022differentially} introduce a method for private adaptive optimization without extra public data. Using delayed preconditioners, they enhance adaptivity and reduce noise. They provide convergence proofs for both convex and non-convex cases. Tests on real-world datasets show improved convergence rates and model utility within a set privacy constraint.

\paragraphb{Adaptive noising} 
A different idea is to optimize the additive noise added to achieve privacy to reduce its deleterious impact on the training process.

Lee et al.~\cite{lee2018concentrated} argue that it is necessary to allocate privacy budget carefully for different iterations. They propose an algorithm (DP-AGD) that adaptively allocates privacy budget for different iterations with zero-mean Concentrated Differential Privacy (CDP)~\cite{bun2016concentrated}.

Xu et al.~\cite{xu2020adaptive} propose a new algorithm called ADADP which implements adaptive learning rate and adaptive noise together to achieve better convergence rate and model performance. To apply adaptive noise, they add noise with smaller variance to the gradients with smaller sensitivity to reduce the effect of noise on the gradient's true direction.

Xiang et al.~\cite{xiang2021differentially} consider the utility subspace of the model. The authors find that if the DP noise is added to some less important directions, they can achieve improved performance. They proposed a directional noise DP mechanism that selects a noise direction with optimal utility.

Zhang et al. \mbox{\cite{zhang2025disk}} introduce DiSK, a framework that enhances DP optimizers by using a simplified Kalman filter to denoise privatized gradients, treating them as noisy observations of a dynamic system to achieve more accurate gradient estimates.

\paragraphb{Improved accounting}
The original moments accountant proposed by Abadi et al.~\cite{abadi2016deep} for DP-SGD has tighter composition bound than advanced composition~\cite{dwork2014algorithmic}. However, when considering variants of the DP-SGD algorithm, privacy accounting can be further enhanced.

Wang et al.~\cite{wang2019efficient} propose a new algorithm called Differentially Private Stochastic Recursive Momentum (DP-SRM) based on a sharp analysis of the privacy guarantee using RDP. They build this algorithm based on the stochastic recursive momentum technique that is able to decrease the variance of gradients. Their algorithm is more scalable and efficient than the original \dpsgd{}. 

Ding et al.~\cite{ding2022differentially} propose Perturbed Iterative Gradient Descent Optimization (PIGDO) which uses gradient descent optimization (GDO) as part of iteration update among some batches, aggregates these batches into a lot and then clips the mean gradients and add noise. By studying the privacy loss of PIGDO, they also propose a new Modified Moments Accountant (MMA) with a tighter bound to save privacy budgets. Their experiments show that PIGDO with MMA outperforms original DP-SGD in terms of utility and privacy.

\paragraphb{Variants of DP-SGD}
In each iteration \dpsgd{} first computes the gradient, clips the gradient, and then adds noise to it. Xiao et al.~\cite{xiao2022differentially} propose a technique called ModelMix. The iterative perturbation in ModelMix has the same gradient computation and clipping as \dpsgd{}, however, before adding Gaussian noise to the clipped gradients, ModelMix performs a random aggregation of the model's weights among the last and penultimate iterations' weights. The authors claim that this random aggregation is equal to adding a randomized proximal term into the objective function and decreasing the noise that is added in the last few iterations.

Xiao et al. \mbox{\cite{xiao2025trustworthy}} introduce Data-Specific Indistinguishability (DSI), a variant of differential privacy that provides trust guarantees by adding optimal, anisotropic Gaussian noise to ensure a model's output is statistically close to outputs from a specific set of safe reference datasets.

Wei et al. \cite{wei2022dpis} present DPIS as an alternative optimizer for DP-SGD. By leveraging importance sampling for mini-batch selection in each SGD iteration, DPIS reduces sampling variance and the necessary random noise for maintaining privacy.

\paragraphb{Speeding up \dpsgd} 
Subramani et al.~\cite{subramani2021enabling} propose a new training method for DP-SGD that combines vectorization, just-in-time compilation, and static graph optimization to reduce running time. They implement their method on JAX and find that it runs $50\times$ faster than the best alternative in some cases.

The most time-consuming and high computation cost part of DP-SGD is per-sample gradient clipping which needs to compute per-example gradients to obtain per-example gradient norm. To overcome this major bottleneck, Ghost clipping~\cite{li2021large,bu2022scalable} is used to compute per-example gradients' norm without actually computing per-example gradients. Essentially, per-example gradient norms can be computed by computing per-example gradient norms for each layer of the model so one only needs to instantiate one per-example gradient tensor. However, this requires two back-propagation passes. To improve from Ghost clipping, Bu et al.~\cite{bu2022differentially} propose a book-keeping technique that could make DP training faster and memory efficient almost as non-DP training. This only requires one back-propagation pass to computes per-example gradient norm like Ghost clipping. 

\paragraphb{Public data} 
Some algorithms assume the existence of public data that can be used by the learning algorithm. Since the public data is by assumption non-sensitive there are no privacy concerns when accessing it. We discuss this assumption further in~\cref{sec:public_data}.

Kairouz et al.~\cite{kairouz2020fast} introduce a differentially private variant of AdaGrad that adds Gaussian noise to both gradients and the preconditioner while projecting updates onto a gradient subspace estimated from public data. They show that, under low-rank subspace structure and decaying gradient norms, this approach achieves faster convergence and dimension-independent excess risk than standard DP-SGD.

Li et al.~\cite{li2022private} propose AdaDPS that uses public data or common knowledge about the private data to estimate gradient statistics to precondition the gradients. Experiments show that their method can improve accuracy on language and image tasks.

\subsection{Teacher-Student Approaches}\label{sec:public_data:teacher}
A different setting using public data is the semi-supervised knowledge transfer setting of Papernot et al.~\cite{papernot2016semi}. In this setting, an unlabeled public dataset is labeled through a differential private process by an ensemble of teacher models trained on disjoint subsets of a sensitive dataset. A student model is trained on the public data augmented with private labels. This technique is called Private Aggregation of Teacher Ensembles (PATE).

Uniyal et al.~\cite{uniyal2021dp} propose that PATE has better performance than \dpsgd{} with the same privacy budget. Papernot et al.~\cite{papernot2018scalable} further improve the technique by proposing a more selective noisy aggregation mechanism. Zhu et al.~\cite{zhu2020private} propose to use KNN for sampling the public data to avoid data splitting in the original PATE setting to improve teacher models' performance. Sun et al.~\cite{sun2020differentially} propose a new voting mechanism with smooth sensitivity called Immutable Noisy ArgMax.

\section{Selected papers \& Methodology} \label{sec:method}

\begin{table}[t!]
\centering
\caption{Selected 11 papers for experiments. Here we use Generalizability (G), and Reliability (R) to measure Reproducibility and Reliability. \CIRCLE{} means the paper satisfies all requirements in that part. \LEFTcircle{} means satisfying part of them and \Circle{} means satisfying none of them. A more detailed version can be found in~\cref{tab:claimed_results}.}
\label{tab:selected_papers}
\resizebox{0.925\columnwidth}{!}{%
\begin{tabular}{lllcc}
\toprule
\multirow{2}{*}{Techniques} & \multirow{2}{*}{Paper} & \multirow{2}{*}{Year} & \multicolumn{2}{c}{Reproducibility} \\
                          &                        &                       & G & R  \\ \midrule
Model architecture        & Klause et al.~\cite{klause2022differentially} & 2022 &  \LEFTcircle   & \Circle   \\
Model architecture        & Sander et al.~\cite{sander2022tan}            & 2023 & \LEFTcircle  & \LEFTcircle      \\
Hyperparameter selection  & Dormann et al. \cite{dormann2021not}          & 2021 & \LEFTcircle & \LEFTcircle     \\
Augmentation multiplicity & De et al.~\cite{de2022unlocking}              & 2022 & \CIRCLE   & \CIRCLE  \\
Augmentation multiplicity & Bao et al.~\cite{bao2024dp}                   & 2024 & \CIRCLE   & \CIRCLE     \\
Feature selection         & Tram\`er and Boneh~\cite{tramer2020differentially} & 2020 & \LEFTcircle    & \LEFTcircle   \\
Gradient Clipping         & Bu et al.~\cite{bu2022scalable}               & 2022 & \LEFTcircle    & \LEFTcircle    \\
Gradient Clipping         & Bu et al. \cite{bu2022automatic}              & 2024 & \CIRCLE   & \LEFTcircle   \\
Fine-tuning technique      & Cattan et al. \cite{cattan2022fine}           & 2022 & \LEFTcircle   & \Circle     \\
Fine-tuning technique      & Luo et al. \cite{luo2021scalable}             & 2021 & \CIRCLE   & \LEFTcircle    \\
Fine-tuning technique      & Tang et al. \cite{tang2023differentially}     & 2024 & \CIRCLE   & \CIRCLE     \\ \bottomrule
\end{tabular}%
}
\end{table}

In this section, we describe how we select papers, our evaluation methodology, research questions, and experimental setup. We then propose a framework to establish whether a method provides statistically significant improvements over a baseline. We justify the need for this framework by empirically measuring the consequences of the additional randomness and variability of DPML (compared to non-private ML).

\subsection{Selection Criteria} \label{sec:selected_papers}
To identify papers for inclusion, we took a deep dive in the DPML literature from 2020 to the time of writing. We also included papers appearing only on arXiv, as some with claimed state-of-the-art methods such as De et al.\mbox{~\cite{de2022unlocking}}, are hosted there.

For a paper to be considered SoTA, it must explicitly claim this status.
To ensure a fair and straightforward comparison, we focused on papers employing DP-SGD methods and presenting their main results on computer vision tasks. 

From our literature search, we identified 11 papers that are the most representative samples, listed in \cref{tab:selected_papers}. These papers were selected based on the following criteria:
\begin{enumerate}
\item They claim to achieve SoTA performance in DPML.
\item Their proposed methods are innovative and straightforward to implement or their code is open source.
\item They represent the latest and most promising research directions, thus providing a comprehensive overview of the current SoTA. 
\end{enumerate}
These characteristics make the selected papers particularly noteworthy and suitable for inclusion.

\subsection{Methodology \& Research questions }
We reproduce each of the 11 selected DPML papers using either code open-sourced by the authors or our re-implementation of techniques as described in the paper. 
We seek to answer the following questions:
\begin{enumerate}[leftmargin=3em,label=(RQ\arabic*),nolistsep,noitemsep]

    \item Do the proposed methods achieve their claims?
    \item Are improvements obtained outside of the experimental settings used in the papers?
    \item What part(s) of the model should be DP fine-tuned?
    \item Can different techniques be combined? 
    \item What techniques are the most promising?    
    \item What are important methodological guidelines to ensure scientifically sound and reliable findings?
\end{enumerate}

\subsection{Experiments Setup} \label{sec:exp_setup} \label{sec:reproducibility:eval}


%


\paragraphb{Datasets}
We chose the following 9 datasets:
CIFAR-10, CIFAR-100, MNIST, Fashion-MNIST, EuroSAT, ISIC 2018, PathMNIST, Caltech 256, SUN397 and Oxford Pet because they are either used in the 11 papers we evaluated or have a large domain gap to the pretrain dataset. We provide more details for these datasets in~\cref{app:dataset}.

\paragraphb{Setup}
To ensure a fair comparison, all our experiments are conducted using PyTorch and Opacus~\cite{opacus}. While some works, such as that of De et al.~\cite{de2022unlocking}, provide open-source code written in JAX, we utilize a reproduced PyTorch version based on the code proposed by Sander et al.~\cite{sander2022tan}. The same codebase was also used by Tang et al.~\cite{tang2023differentially} and Bao et al.~\cite{bao2024dp}. Choice of codebase matters as there are notable performance differences, typically in the range of 1-2\%, between the JAX and PyTorch versions of~\cite{de2022unlocking}, as highlighted in some related work~\cite{sander2022tan,bao2024dp}.


Unless otherwise specified, we adopt the Wide-ResNet 16-4 models as the base model, which is also used in De et al.~\cite{de2022unlocking}, Klause et al.~\cite{klause2022differentially}, Tang et al. \cite{tang2023differentially}, Bao et al. \cite{bao2024dp} and Sander et al.~\cite{sander2022tan}. To implement DP-SGD, we use Opacus~\cite{opacus} and make necessary modifications based on different experimental requirements. All experiments (except pre-trained model experiments) are conducted with the same DP setting --- $\varepsilon = 8$, $\delta = 10^{-5}$, batch size of 4096, clipping bound $C=1$, and 200 training epochs. We report the results for $3$ independent runs.



\subsection{Statistical Framework}\label{sec:methodology:framework}

We propose a framework to enhance our reproducibility experiments and ascertain if a proposed DPML method provides statistically and practically significant improvements.
This framework uses a simplified Cohen's d~\cite{lakens2013calculating} to measure {\em effect size} (improvement size) and paired t-tests~\cite{kim2015t} to measure statistical significance. These methods are well regarded and used across diverse scientific research areas, including psychology~\cite{wetzels2011statistical} and medicine~\cite{austin2008critical}.

Statistical tests come with their own sets of drawbacks and their widespread adoption and (mis)use in some disciplines have led to deleterious practices such as p-hacking~\cite{head2015extent}. Such tests are seldom used in machine learning practice, although there are numerous methods to use them~\cite{grandvalet2006hypothesis,bayle2020cross}. Machine learning research often reports average performance measures such as accuracy (and sometimes also variation or error bars) and relies on this to establish whether one method is superior to another. We argue that for DPML the use of a framework with statistical testing such as the one we propose is warranted. First, we expect larger uncertainty and variability in each measurement (than for non-private ML). This is in part due to inherently greater randomness in DPML as we demonstrate in~\cref{sec:random_seeds}. This is especially striking when the privacy budget is small (e.g., $\varepsilon=0.1$) or the training dataset is small.\footnote{When the training dataset is enormous, as is often the case in deep learning research, statistical measurements may be superfluous since any observed difference is automatically statistically significant. But in such cases, obtaining tight privacy guarantees with performance similar to non-private models may not be difficult in the first place.} Second, obtaining each measurement is often an order of magnitude more computationally onerous for DPML, so relying on only a few measurements may be necessary.

\paragraphb{Framework Details}
Suppose we collect a series of $n$ paired observations from two distinct training runs (e.g., the proposed method vs. the baseline). Let $a_1, a_2, \ldots a_n$ and $b_1, b_2, \ldots b_n$ be the observations for the proposed method and the baseline, respectively. For example, these could be test accuracy measurements on models trained with two different methods but paired such that $a_j$ and $b_j$ are the test accuracies on dataset $j$ (or on run $j$ of $n$ on some fixed dataset). We compute the following:
\smallskip
\begin{itemize}[leftmargin=1.25em]
\item Raw means: $\mu_1 = n^{-1} \sum_{i} a_i$ and $\mu_2 = n^{-1} \sum_{i} b_i$
\item Raw deviations: 
\[
\sigma_1 = \sqrt{\frac{\sum_{i} (a_i - \mu_1)^2}{n}} \quad \text{ and } \quad  \sigma_2 = \sqrt{\frac{\sum_{i} (b_i - \mu_2)^2}{n}}
\]
\item Paired differences: $d_i = a_i - b_i$ for $i=1,2,\ldots,n$.
\item Means and deviation of paired differences: $\mu_d = \sum_i d_i$ and $\sigma_d = \sqrt{\frac{\sum_{i} (d_i - \mu_d)^2}{n}}$. 
\end{itemize}
\smallskip

To conduct the paired t-test which uses the t-distribution \cite{lange1989robust}, the \textit{t-statistic} is calculated as:
$t = \frac{\mu_d}{\sigma_d / \sqrt{n}}$, 
and from it a \textit{p-value} is obtained \cite{goodman2008dirty}. If the p-value is below the set significance level $\alpha$ (e.g., $\alpha=0.05$) this leads to the rejection of the null hypothesis (in our case the two methods perform the same). If the p-value exceeds this threshold, the null hypothesis stands, suggesting the observed difference between the two methods could be due to chance. In our experiments, we use the default parameters of scipy.stats.ttest\_rel to compute the \textit{t-statistic} and p-value.\footnote{\url{https://docs.scipy.org/doc/scipy/reference/generated/scipy.stats.ttest_rel.html}}

We also measure practical significance as {\em improvement size} or {\em effect size} using Cohen's d. That is, we compute: 
\[
d=\frac{{\mu}_1-{\mu}_2}{\sigma_p} \quad \text{ with } \quad \sigma_p = \sqrt{\frac{\sigma_1^2 + \sigma_2^2}{2}} \ .
\] 
where $\mu$ is the mean of each method's result and $\sigma_p$ is the pooled standard deviation~\cite{cohen2013statistical}. Note that the formula for $\sigma_p$ is simplified from the general case, since in our case the two groups always have the same size.

\paragraphb{Test selection and applicability}
We chose to base our framework on the paired t-test and Cohen's d because they are well-known, useful even for small $n$, and straightforward to apply to DPML. 

Using a hypothesis testing procedure is more principled than the oft-used rule-of-thumb of looking for an improvement larger than the standard deviation. The paired test is more powerful than its unpaired test because it reduces variability.\footnote{The paired t-test's statistical power increases with the correlation.} Further, it allows us to compare methods across multiple experimental settings since test accuracy measurements of models trained by each method in a given experimental setting (same model architecture, same dataset, same number of training epochs, etc.) can be paired.

\paragraphb{Test power versus number of runs}
The power of the test depends not only on the number of runs, $n$, but also on the effect size. A statistical rule-of-thumb is that to achieve 80\% or more power for paired t-test at significance level $\alpha=0.05$ requires $n \approx 16 \frac{1}{d^2}$, where $d$ is the effect size (see~\mbox{\cite{van2011statistical}}). This means that when the effect size is large, which is the case experimentally for some of the state-of-the-art DPML methods, only a few runs are needed. However, for methods with small effect sizes, the only way to reliably establish an improvement may be to increase the number of runs until the test becomes sufficiently powerful.

\subsection{Impact of Randomness}\label{sec:random_seeds}
%

%

\begin{table*}[t]
\centering
\caption{Test accuracy on MNIST and CIFAR-10 for 500 runs with different randomly selected seeds. We use WRN-16-4 for training from scratch on MNIST and CIFAR-10, and Vit\_base\_patch16\_224 for fine-tuning on CIFAR-10. Here $\infty$ means using SGD (no privacy). }
\label{tab:random_seeds}
\resizebox{0.9\textwidth}{!}{%
\begin{tabular}{ccccccccc|ccccc}
\toprule
        & \multicolumn{8}{c|}{From scratch}                                                            & \multicolumn{5}{c}{Fine-tuning}                    \\ \cmidrule{2-14} 
Dataset & \multicolumn{4}{c|}{CIFAR-10}                              & \multicolumn{4}{c|}{MNIST}            & \multicolumn{5}{c}{CIFAR-10}                    \\ \midrule
 $\varepsilon$       & $\infty$     & 1   & 4   & \multicolumn{1}{c|}{8}   &   $\infty$     & 1   & 4   & 8   &   $\infty$    & 0.5 & 1   & 4   & 8  \\ \hline
Mean   & 77.56\% & 42.45\% & 60.20\% & \multicolumn{1}{c|}{66.56\%} & 99.17\% & 96.80\% & 98.27\% & 98.51\% & 98.32\% & 96.15\% & 97.83\% & 98.00\% & 98.01\% \\
Std.     & 0.68\%  & 0.81\%  & 0.93\%  & \multicolumn{1}{c|}{0.93\%}  & 0.15\%  & 0.20\%  & 0.28\%  & 0.28\%  & 0.02\%  & 0.23\%  & 0.09\%  & 0.07\%  & 0.05\%  \\
Median & 77.50\% & 42.53\% & 60.22\% & \multicolumn{1}{c|}{66.68\%} & 99.21\% & 96.80\% & 98.37\% & 98.63\% & 98.32\% & 96.14\% & 97.83\% & 98.01\% & 98.02\% \\
Max     & 79.64\% & 45.69\% & 63.47\% & \multicolumn{1}{c|}{68.61\%} & 99.61\% & 97.25\% & 98.67\% & 98.93\% & 98.33\% & 96.61\% & 98.06\% & 98.12\% & 98.14\% \\
Min     & 76.03\% & 40.20\% & 57.90\% & \multicolumn{1}{c|}{64.13\%} & 98.68\% & 95.60\% & 96.60\% & 97.09\% & 98.17\% & 95.39\% & 97.45\% & 97.79\% & 97.86\% \\
Max-Min & 3.61\%  & 5.49\%  & 5.57\%  & \multicolumn{1}{c|}{4.49\%}  & 0.93\%  & 1.65\%  & 2.07\%  & 1.84\%  & 0.16\%  & 1.22\%  & 0.61\%  & 0.33\%  & 0.28\%  \\ \bottomrule
\end{tabular}%
}
\end{table*}

Randomness in the initial weights and mini-batch sampling steps during the training process (e.g., with SGD) plays a significant role in machine learning. However, for DPML, randomness due to the added noise to achieve differential privacy also impedes convergence and decreases the performance of the trained model. Moreover, this additional randomness can obscure the relationship between alternative methods' performance because it tends to increase variability across runs.

We study this phenomenon through a set of principled experiments where we train models with SGD and \dpsgd{} while varying the random seeds used by the random number generator, thereby allowing us to observe the impact of randomness on model performance. Here it is worth pointing to related work by Picard~\cite{picard2021torch} studying the impact of random seeds on computer vision.

We trained a WRN-16-4 on CIFAR-10 from scratch using 500 randomly chosen seeds. Similarly, we fine-tuned a Vit-base-patch16-224 on CIFAR-10 with 500 randomly chosen seeds. The mean, median, max, min test accuracies, stddev, and the difference between max and min test accuracies from the 500 runs for different $\varepsilon$ values and non-private cases. Results are presented in~\cref{tab:random_seeds}.

Observe that both the standard deviation and max-min difference is often much greater for \dpsgd{} than SGD (represented as $\varepsilon=\infty$ in~\cref{tab:random_seeds}) e.g. \textbf{5.57\%} vs 3.61\%. In other words, the noise added to the gradient increases the variability in the quality of found solutions. This is in some sense expected given the privacy constraint. However, it suggests that for DPML compared to non-private ML: (1) a larger performance gap between two competing methods is needed to conclude that one outperforms another, and (2) there is greater potential for ``{\em seed hacking},'' where unethical researchers specifically select seeds to unfairly claim an advantage for their methods. We discuss this in~\cref{sec:exp:seeds_hacking}.

Another relevant observation from~\cref{tab:random_seeds} is that the standard deviation and max-min difference are much lower for fine-tuning compared to the train-from-scratch setting. This suggests that a promising methodological step to ensure reproducibility is to evaluate a method both from scratch and fine-tuning settings. We also included MNIST models trained from scratch in the experiments to establish that variability in the fine-tuning setting is in fact lower and that the observed results are likely not due to this setting typically yielding higher accuracy models.

Finally, we observe higher variability at lower privacy budgets, which is expected but underlines the importance of evaluating methods in a wide range of privacy regimes.



\subsection{Seed Hacking?}\label{sec:exp:seeds_hacking}

In this section, we explore the concept of {\em seed hacking}, inspired by the work of Picard~\cite{picard2021torch}. Seed hacking refers to the process of selectively choosing a small subset of random seeds that anomalously enhance the performance of a proposed method over a baseline, giving a false impression of improvement when, in fact, no genuine enhancement exists from the proposed technique~\cite{gardner2018enabling,henderson2018deep}. Note that we are not accusing anyone of engaging in seed hacking; rather, we bring up this possibility as a loose analog to the problem of $p$-hacking~\cite{head2015extent} but also to emphasize the significance of the increased randomness in DPML.

We use data from our random seed experiment (\cref{tab:random_seeds}) to simulate a seed-hacking strategy on the fine-tuning task using the CIFAR-10 dataset. For each privacy budget value, we randomly select 10 out of 500 experimental runs, rank these by performance, and choose the top three outcomes for the ``proposed method'' to simulate what an unethical researcher engaging in seed hacking might do. For the ``baseline method'', by contrast, we randomly select three outcomes from the same 500 runs.

We conduct two types of statistical tests. The first test employs a common approach where if the mean performance of the proposed method minus its standard deviation exceeds the baseline’s mean performance plus its standard deviation, the proposed method is considered superior. The second test applies our proposed framework to conduct paired t-tests with significance denoted by a p-value less than $\alpha = 0.05$. The experiment simulates seed hacking 1000 times and each case executes both tests. 

Results are shown in~\cref{tab:seeds_hacking}, where each entry is the number of passing instances of the test. Observe that non-private ML (SGD --- $\varepsilon = \infty$) has the lowest numbers. The numbers are greater in the DPML case, especially for low privacy budgets. More importantly, the std test has a much higher false discovery rate than the paired t-test. Note that if seeds were selected randomly instead (no seed hacking) then we would expect an average $50$ passing instances in each cell, reflecting a false discovery rate of $\alpha=0.05$.

The table reports only rates of false discoveries (Type I errors) because the experiment simulates a scenario where a researcher cherry-picks seeds. With all results coming from the same training method, there are no notable improvements (the Null hypothesis is true by definition) and so there are no Type II errors.

\begin{table}[t]
\centering
\caption{Number of times that seeds hacking show proposed method better than baseline among 1000 independent runs.}
\label{tab:seeds_hacking}
\resizebox{0.65\columnwidth}{!}{%
\begin{tabular}{cccccc}
\toprule
$\varepsilon$ & $\infty$ & 0.5 & 1   & 4   & 8   \\ \midrule
std test                   & 345                 & 402 & 397 & 382 & 357 \\
t-test                     & 141                 & 191 & 186 & 174 & 153
\\ \bottomrule
\end{tabular}%
\vspace{-5pt}
}
\end{table}

\paragraphb{Random seeds and differential privacy}
There is another subtle but critical difference in the role that random seeds play in DPML versus non-private ML. The reader may wonder why it is unacceptable to optimize the choice of random seed --- putting aside for a moment the dishonesty related to seed hacking. After all, if some choices of random seeds are better than others, why not pick the seed that yields the best model? Arguably what matters most is the model that is actually used, not the distribution of possible models that we could have trained. Further, from a pure reproducibility standpoint, fixing the seed to a known value to eliminate its impact is desirable.

This reasoning does {\em not} run through for DPML because the randomness of the seed is required for the differential privacy guarantee. Since the random seed determines the added noise to the gradient in \dpsgd, fixing it or selecting it on any criteria that an adversary knows about (or can replicate) reduces the uncertainty about the noise distribution, which thus breaks the privacy guarantee. Therefore the seed must truly be selected at random.





\section{R+R Experiments} \label{sec:exp:train_from_scratch}

In this section, we conduct our R+R (Reproducibility+Replicability) evaluations on our 11 selected papers. We not only attempt to reproduce their results and check if they match their claims in their papers. We also expand the evaluation settings (e.g., to new datasets, or new model architectures) to ascertain whether the proposed methods still deliver improvements over baselines.

\begin{table}[t]
\centering
\caption{Reproduced test accuracy for Dormann et al. \cite{dormann2021not}}
\label{tab:hyper-parameter_results}
\resizebox{0.9\columnwidth}{!}{%
\begin{tabular}{cccc}
\toprule
$\varepsilon$                          & 1.93            & 4.21            & 7.42           \\ \midrule
Claimed & 58.6\% (0.38\%)  & 66.2\% (0.38\%)  & 70.1\% (0.20\%) \\
Reproduced & 58.64\% (1.16\%) & 66.41\% (0.78\%) & 68.88\% (1.62\%)  \\ \bottomrule
\end{tabular}%
}
\end{table}
\begin{table}[t]

\caption{Sander et al. \cite{sander2022tan} performance for varying $\varepsilon$.}
\label{tab:order_eps}
\resizebox{\columnwidth}{!}{%
\begin{tabular}{lcccc}
\toprule
Order & 0 & 1 & 2 & 3 \\ \midrule
$\varepsilon=8$ & 71.68\% ($\pm0.50\%$) & 65.74\% ($\pm0.40\%$) & 66.85\% ($\pm1.60\%$) & \textbf{74.07\% ($\pm0.40\%$)} \\ 
$\varepsilon=1$ & 52.75\% ($\pm0.22\%$) & 43.10\% ($\pm0.51\%$) & 44.71\% ($\pm0.65\%$) & \textbf{52.96\% ($\pm0.32\%$) }\\ 
$\varepsilon=0.5$ & 47.00\% ($\pm0.54\%$) & 40.51\% ($\pm0.23\%$) & 42.54\% ($\pm0.54\%$) & \textbf{47.51\% ($\pm0.36\%$)} \\ 
$\varepsilon=0.1$  & 32.19\% ($\pm1.24\%$) & 25.64\% ($\pm1.15\%$) & 29.53\% ($\pm0.85\%$) & \textbf{32.57\% ($\pm0.92\%$)} \\ \bottomrule
\end{tabular}%
}
\end{table}

%
\paragraphbe{Dormann et al. \cite{dormann2021not} --- Hyperparameter selection.}
Dormann et al.~\cite{dormann2021not} focuses on hyperparameter selection, advocating for the adoption of large batch sizes (high sampling rate) combined with a higher noise level. 

We used their official codebase to reproduce their experiments on CIFAR-10~(\cref{tab:hyper-parameter_results}).\footnote{\url{https://github.com/OsvaldFrisk/dp-not-all-noise-is-equal}} Our results are consistent with their assertions. Employing larger batch sizes has also been endorsed and adopted by more recent studies such as De et al.~\cite{de2022unlocking} and Bu et al.~\cite{bu2022scalable} which shows replicability. Consequently, we use this hyperparameter selection strategy as default in subsequent experiments.

\takeaway{Large batch sizes and high noise levels consistently provide superior performance in experiments across a wide variety of scenarios.}

\begin{table*}[th]
\centering
\caption{Average test accuracy ($\pm$ standard deviation) for reproducing SoTA methods. The privacy budget for training is 8 and $\delta=10^{-5}$. We use the same DP setting for all experiments, i.e., batch size is 4096, $C=1$, 200 training epochs.}
\label{tab:ex_results}
\resizebox{1\textwidth}{!}{%
\begin{tabular}{ccccc}
\toprule
Order                       & 0              & 1              & 2              & 3                       \\ \midrule
Baseline (WRN-16-4)          & 71.68\% ($\pm0.50\%$) & 65.74\% ($\pm0.40\%$) & 66.85\% ($\pm1.60\%$) & \textbf{74.07\% ($\pm0.40\%$)} \\ 
Baseline + ScaleNorm        & 72.71\% ($\pm1.20\%$) & 63.87\% ($\pm1.40\%$) & 64.18\% ($\pm2.90\%$) & \textbf{72.95\% ($\pm0.60\%$)}          \\ 
Baseline + Mixed Ghost Clipping & \textbf{72.96\% ($\pm0.30\%$)} & 66.91\% ($\pm1.50\%$) & 67.26\% ($\pm2.30\%$) & 72.79\% ($\pm0.60\%$) \\ 
Baseline + Self-augmentation~\cite{de2022unlocking}& 77.79\% ($\pm0.50\%$) & 68.87\% ($\pm0.80\%$) & 69.68\% ($\pm0.90\%$) & \textbf{78.10\% ($\pm0.50\%$)}          \\ 
Baseline + \selfmix~\cite{bao2024dp} & 78.49\% ($\pm0.21\%$) & 69.17\% ($\pm0.45\%$) & 69.89\% ($\pm0.42\%$) & \textbf{79.83\% ($\pm0.32\%$)}          \\ 
Baseline + Self-augmentation~\cite{de2022unlocking} + ScaleNorm & 77.43\% ($\pm0.30\%$) & 66.40\% ($\pm0.9\%$) & 67.37\% ($\pm0.40\%$) & \textbf{78.19\% ($\pm0.20\%$)}          \\ \bottomrule
\end{tabular}%
}
\end{table*}
%

\paragraphbe{Sander et al.~\cite{sander2022tan} --- Changing order.}
Recall that Sander et al.~\cite{sander2022tan} proposed changing the order of activation function and normalization layers to obtain a performance boost of 5\% to 10\%.\footnote{Although Sander et al.~\cite{sander2022tan} proposed other improvements, we are primarily interested in evaluating their changing order method.} As a baseline, we train the WRN-16-4 model (on \cifar{}) using \dpsgd{} to achieve an average test accuracy of $71.68\%$ (\cref{tab:ex_results}).

We empirically explored four different ordering schemes as represented in their code.\footnote{Order 0, where the order of layers follows Conv-ReLU-GN, with the same order in the shortcut (here Conv means convolution layers, ReLU is the activation function and GN means group normalization layer); Order 1, where the order of layers follows Conv-GN-ReLU, with the same order in the shortcut; Order 2, where the order of layers follows Conv-GN-ReLU, but the shortcut follows Conv-ReLU-GN; and Order 3, where the order of layers follows Conv-ReLU-GN, but the shortcut follows Conv-GN-ReLU. Results applied to the WRN-16-4 model} Results are shown in~\cref{tab:ex_results}. The results are consistent with the claims made by Sander et al.~\cite{sander2022tan} as we observed a 5\% to 10\% boost in performance. Specifically, using order 3 resulted in the best performance, with an average test accuracy of $74.07\%$. 

Although Sander et al.~\cite{sander2022tan} only report results for $\varepsilon = 8$, we perform experiments varying $\varepsilon$ from $\varepsilon=0.1$ to $\varepsilon=8$ and show results in~\cref{tab:order_eps}. We find that the same pattern holds across $\varepsilon$ values, showing that changing layer order as specified in their paper does indeed replicate.

\paragraphb{De et al.~\cite{de2022unlocking} --- Self-augmentation}
De et al.~\cite{de2022unlocking} achieved a new SoTA performance on \cifar{} through hyperparameter tuning and a combination of techniques such as self-augmentation (aka augmentation multiplicity), weight standardization, and parameter averaging (ema). They use a codebase based on JAX, whereas we use a PyTorch version of it from Sander et al. \cite{sander2022tan}.\footnote{\url{https://github.com/facebookresearch/tan}}

For reproducibility, the results we obtain are shown in~\cref{tab:ex_results}. The method significantly improves the model's performance (from 71.68\% to 77.79\%), which is consistent with De et al.'s claims. Although they reported slightly higher performance, we believe that our results are comparable, considering the randomness in training and the fact that we use a different codebase.

To evaluate replicability, we extend their proposed method by testing it on a wider range of $\varepsilon$ from 0.1 to 8 on CIFAR-10 and present the results in \cref{tab:various_eps}. We observe that their proposed improvements decrease as the privacy budget is decreased. When $\varepsilon = 0.1$, the improvement is marginal. This is in stark contrast to the result obtained varying $\varepsilon$ for order switching (Sander et al.~\cite{sander2022tan}).

\begin{table}[!]
\caption{We fine-tune Clip-Vit-B-16 models on Caltech256, SUN397 and Oxford Pet datasets using different $\varepsilon$ with $\delta=10^{-5}$, and report the test accuracy (\%). We can observe that \selfmix and \diffmix, outperform the baselines in all cases.}
\label{tab:dp_mix_fine-tune}
\centering
\resizebox{1\linewidth}{!}{%
\begin{tabular}{cc|cccc}
\toprule
            \textbf{Dataset}                    & \textbf{Method}               & $\mathbf{\varepsilon=1}$            & $\mathbf{\varepsilon=2}$           & $\mathbf{\varepsilon=4}$            & $\mathbf{\varepsilon=8}$         \\ \midrule
                           
\multirow{4}{*}{Caltech256} 
& Self-Aug~\cite{de2022unlocking}    & 80.36($\pm.11$) & 89.67($\pm.16$) & 92.01($\pm.08$) & 93.17($\pm.15$) \\
& \selfmix   & 81.21($\pm.15$) & 90.12($\pm.17$) & 92.17($\pm.21$) & 93.39($\pm.08$) \\
& \diffmix   & \textbf{89.69($\pm.23$)} & \textbf{91.82($\pm.15$)} & \textbf{92.86($\pm.14$)} & \textbf{93.87($\pm.10$)} \\ 
\midrule
\multirow{4}{*}{SUN397} 
& Self-Aug~\cite{de2022unlocking}    & 72.65($\pm.09$) & 76.02($\pm.14$) & 78.05($\pm.11$) & 79.54($\pm.15$) \\
& \selfmix   & 73.19($\pm.13$) & 76.45($\pm.17$) & 78.67($\pm.16$) & 79.57($\pm.14$) \\
& \diffmix   & \textbf{75.12($\pm.17$)} & \textbf{77.78($\pm.12$)} & \textbf{79.47($\pm.18$)} & \textbf{80.57($\pm.09$)} \\ \midrule 
%
\multirow{4}{*}{Oxford Pet} & Self-Aug~\cite{de2022unlocking}    & 72.21($\pm.21$) & 82.11($\pm.19$) & 85.84($\pm.25$) & 88.23($\pm.11$) \\
& \selfmix   & 72.45($\pm.24$) & 82.51($\pm.21$) & 86.75($\pm.17$) & 88.70($\pm.15$) \\
& \diffmix   & \textbf{83.24($\pm.26$)} & \textbf{86.28($\pm.19$)} & \textbf{88.25($\pm.24$)} & \textbf{89.41($\pm.21$)} \\ 
%
\bottomrule
\end{tabular}%
}
\end{table}
\paragraphb{Bao et al. \cite{bao2024dp} --- DP-Mix}
Bao et al. \cite{bao2024dp} proposed two new techniques \selfmix and \diffmix which achieve SoTA results on multiple datasets. We reproduce their results using their official codebase.\footnote{\url{https://github.com/wenxuan-Bao/DP-Mix}} 
The reproducible results of \selfmix we obtain are shown in~\cref{tab:ex_results}. The method significantly improves the model's performance (from 71.68\% to 79.83\% for baseline and from 78.10\% to 79.83\% for De et al. \cite{de2022unlocking}), which is consistent with Bao et al.'s claims. We also reproduce \diffmix using the same settings as the authors presented and show results in~\cref{tab:dp_mix_fine-tune}. We observe that \selfmix and \diffmix improve the test accuracy in all cases, especially for Caltech256 and Oxford Pet datasets. 

We also extended their method to a wider range of privacy budgets (i.e., $\varepsilon$ ranging from 0.1 to 8) on CIFAR-10 and show the results in~\cref{tab:various_eps}. We observe that the performance boost from the method decreases significantly as the privacy budget decreases, similar to the method proposed by De et al.\cite{de2022unlocking}.

\takeaway{Augmentation multiplicity delivers remarkable performance improvements in practice. It also appears to be a promising direction for future research, albeit its applicability beyond computer vision remains unclear.}

\begin{table}[t]
\caption{De et al.~\cite{de2022unlocking} and Bao et al. ~\cite{bao2024dp} proposed method performance on CIFAR-10  under different $\varepsilon$. Results show that the impact of the improvements decreases as $\varepsilon$ decreases.}

\label{tab:various_eps}
\resizebox{\columnwidth}{!}{%
\begin{tabular}{ccccc}
\toprule
$\varepsilon$                & 0.1             & 0.5             & 1               & 8               \\ \midrule
Baseline            & 32.19\%($\pm1.24\%$) & 47.00\%($\pm0.54\%$) & 52.75\%($\pm0.52\%$) & 71.68\%($\pm0.55\%$) \\ 
De et al. \cite{de2022unlocking} & 32.42\%($\pm1.03\%$) & 48.98\%($\pm0.42\%$) & 56.06\%($\pm0.45\%$) & 77.79\%($\pm0.50\%$) \\ 

Bao et al. \cite{bao2024dp} & \textbf{32.57\%($\pm1.23\%$)} & \textbf{49.14\%($\pm0.49\%$)} & \textbf{57.24\%($\pm0.42\%$)} & \textbf{78.49\% ($\pm0.21\%$)} \\

\bottomrule
\end{tabular}%
}

\end{table}

\paragraphbe{Klause et al.~\cite{klause2022differentially} --- ScaleNorm.}
Recall that Klause et al.~\cite{klause2022differentially} proposed to add a normalization layer (ScaleNorm) after the residual block to achieve better performance.

We added Scale Normalization layers to the model and evaluated the performance of the modified model on \cifar{} (\cref{tab:ex_results}). We find that this improves the model's performance by about 1\%, which is consistent with the results reported in the paper. However, when we applied Sander et al.'s method of changing the order of the activation function and normalization layer, the performance of the model decreased. This result suggests that the benefits of ScaleNorm may not be widely applicable and (or) may not be combined with other techniques.

\paragraphbe{Bu et al.~ \cite{bu2022scalable} --- Mixed ghost clipping.}
Recall that Bu et al.~\cite{bu2022scalable} propose a gradient clipping method using pre-trained Transformer models. However, without experiments on models trained from scratch or ablation studies, it is not clear whether the observed improvements are due to the clipping technique, the pre-trained models, or both.

We apply Mixed Ghost Clipping to the WRN-16-4 model and train it from scratch using an implementation based on the code provided by the authors.\footnote{\url{https://github.com/woodyx218/private_vision}} We find that the method only slightly improved performance~(\cref{tab:ex_results}).

\begin{table}[t]
\centering
\caption{Results of pre-trained transformers for independent $3$ runs. The privacy budget $\varepsilon = 1$ and $\delta = 10^{-5}$. We set the training epochs to 2 and batch size of 5000 as Bu et al.\cite{bu2022scalable} suggested. } 

\label{tab:pretrain_results}
\resizebox{\columnwidth}{!}{%
\begin{tabular}{cccccc}
\toprule
{\color[HTML]{333333} Model} & {\color[HTML]{333333} Clip method} & {\color[HTML]{333333} Test accuracy} & $t$ & p-value & Effect size  \\ \midrule
                                          & Basic clipping       & 94.71\%($\pm0.14\%$)          \\  
\multirow{-2}{*}{CrossViT\_base\_224}     & Mixed Ghost clipping & \textbf{94.77\%($\pm0.13\%$)} &\multirow{-2}{*}{0.78} &\multirow{-2}{*}{0.52}  &\multirow{-2}{*}{0.44} \\ 
\midrule
                                          & Basic clipping       & \textbf{97.34\%($\pm0.20\%$)} \\  
\multirow{-2}{*}{Vit\_base\_patch16\_224} & Mixed Ghost clipping & 95.05\%($\pm0.13\%$)   &\multirow{-2}{*}{-16.61} &\multirow{-2}{*}{\textbf{0.004}}  &\multirow{-2}{*}{-13.57}        \\ \bottomrule
\end{tabular}%
}
\end{table}

We also tested the performance using pre-trained Transformers, following the same setting as the paper. We used the CrossVit-base-224 and Vit-base-patch16-224 models, pre-trained on ImageNet and provided by timm\footnote{\url{https://github.com/huggingface/pytorch-image-models}}, and fine-tuned the models on \cifar{}. We compared Mixed Ghost clipping and basic gradient clipping, with $\varepsilon = 1$ and two training epochs, as suggested in the paper (\cref{tab:pretrain_results}). Mixed Ghost Clipping \textit{did not} outperform basic gradient clipping in a statistically significant way (\cref{sec:methodology:framework}).
When the model architecture is Vit-base-patch16-224, basic gradient clipping achieves better performance (in this case the effect size is massive and the result is statistically significant). Moreover, we did not observe the out-of-memory problem reported in the original paper when using basic gradient clipping.

While running these experiments we inadvertently achieved a new SoTA performance of \textbf{97.34\%} for fine-tuning pre-trained models on \cifar{} with a privacy budget of $\varepsilon = 1$. In this case, the SoTA result was achieved (accidentally) by using a powerful model architecture and tuning some hyperparameters and we argue it does not constitute a meaningful improvement. 

\takeaway{Differences between claimed SoTA results and baselines are sometimes so small that one may accidentally achieve new SoTA results. Such small differences may also not be statistically significant. This highlights the risk with chasing SoTA performance as a strategy for DPML research. Arguably, researchers should focus on designing novel techniques that have a meaningful rationale or are otherwise expected to be reliable and generalizable.}

\begin{table}[!]
\centering
\caption{Reproduced results for Bu et al. \cite{bu2022automatic}. According to the framework of~\cref{sec:methodology:framework}, both clipping methods achieve similar performance, with neither being statistically superior. }
\label{tab:auto_results}
\resizebox{1\columnwidth}{!}{%
\begin{tabular}{cccccc}
\toprule
Dataset & Abadi's clipping       & Auto clipping  & $t$ & p-value & Effect size           \\ \midrule
MNIST   & 98.15\% ($\pm$0.14\%) & 98.14\% ($\pm$0.12\%)  &-0.43 &0.71 &-0.22 \\
Fashion MNIST  & 86.65\% ($\pm$0.35\%) & 86.72\% ($\pm$0.32\%)  &0.37&0.75 &0.21 \\ \bottomrule
\end{tabular}%
}
\end{table}
\paragraphbe{Bu et al. \cite{bu2022automatic} --- Auto clipping.}
Recall that Bu et al. \cite{bu2022automatic} introduced an alternative gradient clipping technique termed ``Auto Clipping.'' As recommended in their paper, we modified the Opacus library to incorporate this new clipping method and conducted experiments on MNIST and Fashion MNIST. \cref{tab:auto_results} shows our results, which are consistent with the performance reported in Bu et al.~\cite{bu2022automatic}. In their paper, a stated goal is to improve the hyperparameters search time of \dpsgd{} by reducing the number of hyperparameters that need to be tuned, which they achieve. However, their method does not outperform the original clipping proposed by Abadi et al.~\cite{abadi2016deep}. According to our framework, neither method can be claimed to provide superior performance with statistical significance.

\takeaway{Claimed improvements of methods may not always achieve statistical significance. Our proposed framework or other statistically valid methods should be used to conclusively determine whether a method truly outperforms its baseline.}

\paragraphbe{Cattan et al.~\cite{cattan2022fine} --- First \& last fine-tuning.}
Cattan et al. \cite{cattan2022fine} argue that fine-tuning the first and last layers yields superior results to only fine-tuning the last layer or the entire model. However, in our experiments, we observed the opposite in \cref{tab:finetune_results}: fine-tuning the first and last layers yields similar or worse performance compared to just fine-tuning the last layer or even the whole model in some cases. 

A plausible explanation for this discrepancy is that the claims by Cattan et al. are limited to their settings, such as using ResNet on CIFAR-10 and CIFAR-100. For our experiments, we use {Vit\_base\_patch16\_224} pre-trained on ImageNet and test their method using multiple different datasets including CIFAR-10, EuroSAT, ISIC 2018, Caltech256, SUN397, and Oxford Pet.

\paragraphbe{Luo et al.~\cite{luo2021scalable} --- Sparse fine-tuning.}
Luo et al.~\cite{luo2021scalable} propose fine-tuning the classification layer, normalization layer, and a minor subset (i.e., 1\%) of the convolution layer parameters. Our replication corroborates that this fine-tuning approach outperforms fine-tuning the entire model (baseline), across all three datasets they used in their paper. However, we find that the proposed 1\% parameter selection is not universally optimal. For instance, selecting 10\% of parameters provides comparatively superior performance. We also find that their proposed method does not achieve better performance compared to only fine-tuning the last layer for datasets such as Caltech-256, SUN 397 and Oxford Pet which are not tested by the authors.

\paragraphbe{Tram\`er and Boneh~\cite{tramer2020differentially} --- Hand-crafted features.}
\begin{table}[t]
\centering
\caption{Our reproduced results for Tram\`er and Boneh~\cite{tramer2020differentially} and compare it to De et al. \cite{de2022unlocking} for $\varepsilon$ from 1 to 8. We find that Tram\`er and Boneh~\cite{tramer2020differentially}  excel in performance with a limited privacy budget ($\varepsilon=1$ and $\varepsilon=2$). However, when the privacy budget exceeds 4, their performance plateaus. (For statistical tests, we take Tram\`er and Boneh~\cite{tramer2020differentially}  as proposed method and De et al. \cite{de2022unlocking} as baseline.)}
\vspace{0.5cm}
\label{tab:hand-crafted}
\resizebox{1\columnwidth}{!}{%
\begin{tabular}{lllll}
\toprule
Paper                  & $\varepsilon$=1                 & $\varepsilon$=2                 & $\varepsilon$=4        & $\varepsilon$=8                 \\ \midrule
\cite{de2022unlocking}      & 56.80\%($\pm$0.49\%)          & 62.90\%($\pm$0.32\%)          & 69.45\%($\pm$0.41\%) & \textbf{78.74\%($\pm$0.45\%)} \\
\cite{tramer2020differentially}  & \textbf{60.88\%($\pm$0.33\%)} & \textbf{66.96\%($\pm$0.56\%)} & 69.74\%($\pm$0.24\%) & 72.40\%($\pm$0.11\%)          \\
\midrule
$t$-statistic & 21.25 & 13.35 & 1.67 & -60.93 \\
p-value & \textbf{0.002} & \textbf{0.006}  & 0.238 & \textbf{0.003}  \\
Effect size & 9.77 & 8.90 & 0.86 & -19.36 \\
\bottomrule
\end{tabular}
}
\end{table}
We reproduced Tram\`er and Boneh~\cite{tramer2020differentially} using their official code and their best method (ScatterNet+CNN) to train a model from scratch on CIFAR-10.\footnote{\url{https://github.com/ftramer/Handcrafted-DP}} We then compare the results to De et al.~\cite{de2022unlocking} under varying privacy budgets (\cref{tab:hand-crafted}). Tram\`er and Boneh~\cite{tramer2020differentially} outperform De et al. \cite{de2022unlocking} when the privacy budget is limited. However, for $\varepsilon = 8$ (or larger), De et al. \cite{de2022unlocking} provide substantially better performance. Further experiments with an increasing privacy budget show performance plateauing for Tram\`er and Boneh whereas the test accuracy for De et al. keeps increasing. 

\takeaway{It is not uncommon for one method to outperform another in one privacy regime but have the reverse occur in a different privacy regime. This underlines the importance of reporting results with a wide range of~$\varepsilon$ values to ensure comprehensive comparisons between methods.}

\begin{table}[tb]
\centering
\caption{Our reproduced results on CIFAR-10 and PathMNIST for Tang et al. \cite{tang2023differentially} using different datasets for Phase 1 with $\varepsilon = 1$.}
\label{tab:random_process}
\resizebox{1\columnwidth}{!}{%
\begin{tabular}{lcc}
\toprule
Method                         & CIFAR-10 & PathMNIST \\ \midrule
Phase1 w/ Random processes & 72.48\% ($\pm0.21\%$)  & 90.65\% ($\pm0.18\%$)  \\
Phase1 w/ EuroSAT             & 70.81\% ($\pm0.32\%$) & 90.58\% ($\pm0.25\%$)   \\ \bottomrule
\end{tabular}
}
\end{table}
\begin{table}[!]
\centering
\caption{FID values between pre-training data (random process data and fine-tuning set --- CIFAR-10 and PathMNIST)}
\label{tab:fid_random_process}
\resizebox{0.7\columnwidth}{!}{%
\begin{tabular}{ccc}
\toprule
   Dataset                  & Random processes & EuroSAT \\ \midrule
CIFAR-10 Train & 107.78              & 123.33  \\
CIFAR-10 Test     & 158.69              & 160.48  \\
PathMNIST Train & 196.61              & 201.71  \\
PathMNIST Test     & 196.75              & 202.64  \\ \bottomrule
\end{tabular}%
}
\end{table}
\paragraphbe{Tang et al.~\cite{tang2023differentially} --- Random-process pretraining.}
Tang et al.~\cite{tang2023differentially} advocate for initially pre-training a model on data produced from random processes, and then fine-tuning it using the private dataset. We replicated their method using their official codebase and validated their performance as shown in~\cref{tab:random_process}.\footnote{\url{https://github.com/inspire-group/DP-RandP}}

Their method essentially initializes the model using random process data instead of public pre-training data. Model performance typically improves with the similarity between pre-training and fine-tuning datasets. In practice, dissimilar datasets to the fine-tuning data are more likely to be publicly available. So this approach raises the question of whether random process data is always beneficial. We evaluate this using the Fréchet Inception Distance (FID)~\cite{heusel2017fid} as a measure of the domain gap.

As shown in~\cref{tab:random_process}, the performance obtained when pre-training on EuroSAT is analogous to that achieved with random process data. The FID values between private datasets (e.g., CIFAR-10 and PathMNIST) and public datasets (e.g., random process data and EuroSAT), are shown in appendix~\cref{tab:fid_random_process}. Note that we selected these datasets specifically to have large FID values. Interestingly, despite EuroSAT exhibiting a larger domain gap compared to random process data, models pre-trained on EuroSAT still deliver performance metrics that closely align with those trained using random process data.

\begin{table*}[t]
\centering
\caption{Test accuracy using Vit\_base\_patch16\_224 on CIFAR-10, EuroSAT, and ISIC 2018 using different fine-tuning methods.}
\label{tab:finetune_results}
\resizebox{1.0\textwidth}{!}{%
\begin{tabular}{cccccccccccccc}
\toprule
 &
  \multicolumn{4}{c}{} &
  \multicolumn{3}{c}{Luo et al. \cite{luo2021scalable}} &
  \multicolumn{3}{c}{Random Subset} &
  \multicolumn{3}{c}{Partial Training} \\ \cmidrule{6-14}
Dataset &
  Whole model &
  First and last \cite{cattan2022fine} &
  Last only &
  Non-private &
  1\% &
  2\% &
  10\% &
  1\% &
  2\% &
  10\% &
  2 &
  3 &
  6 \\ \midrule
CIFAR-10 &
  97.82\%(0.08\%) &
  95.41\%(0.14\%) &
  95.68\%(0.12\%) &
  98.34\%(0.04\%) &
  97.85\%(0.06\%) &
  97.86\%(0.09\%) &
  \textbf{97.93\%(0.11\%)} &
  97.89\%(0.06\%) &
  97.92\%(0.08\%) &
  97.91\%(0.08\%) &
  97.91\%(0.12\%) &
  97.90\%(0.10\%) &
  97.85\%(0.08\%) \\
EuroSAT &
  95.75\%(0.44\%) &
  93.74\%(0.17\%) &
  94.13\%(0.15\%) &
  98.69\%(0.07\%) &
  95.78\%(0.21\%) &
  95.23\%(0.75\%) &
  95.87\%(0.22\%) &
  95.67\%(0.72\%) &
  95.94\%(0.26\%) &
  \textbf{96.15\%(0.30\%)} &
  95.74\%(0. 13\%) &
  95.78\%(0.13\%) &
  95.29\%(0.56\%) \\
ISIC 2018 &
  72.34\%(0.16\%) &
  67.78\%(0.27\%) &
  67.76\%(0.31\%) &
  90.58\%(0.49\%) &
  71.78\%(0.58\%) &
  71.79\%(0.83\%) &
  \textbf{72.51\%(0.46\%)} &
  72.15\%(0.41\%) &
  71.73\%(0.59\%) &
  72.41\%(0.35\%) &
  72.31\%(0.57\%) &
  70.82\%(0.84\%) &
  71.44\%(0.38\%) \\ 
  Caltech 256 &
  30.55\%(0.19\%) &
  80.58\%(0.14\%) &
  \textbf{80.74\%(0.15\%)} &
  95.61\%(0.14\%) &
   42.56\%(0.25\%)&
   31.05\%(0.29\%)&
   30.37\%(0.28\%)&
  31.63\%(0.23\%) &
  29.56\%(0.31\%) &
  29.53\%(0.26\%) &
  31.61\%(0.36\%) &
  29.46\%(0.34\%) &
  30.65\%(0.39\%) \\ 
  SUN 397 &
  43.53\%(0.24\%) &
  67.78\%(0.29\%) &
  \textbf{67.90\%(0.24\%)} &
  84.49\%(0.34\%) &
   42.93\%(0.35\%)&
   43.51\%(0.37\%)&
   43.96\%(0.32\%)&
  43.42\%(0.36\%) &
  44.12\%(0.43\%) &
  43.56\%(0.37\%) &
  43.47\%(0.39\%) &
  42.38\%(0.44\%) &
  42.74\%(0.31\%) \\
  Oxford Pet &
  34.81\%(0.34\%) &
  73.80\%(0.21\%) &
  \textbf{73.92\%(0.35\%)} &
  92.89\%(0.19\%) &
   42.30\%(0.39\%)&
   39.62\%(0.41\%)&
   36.48\%(0.37\%)&
  35.79\%(0.34\%) &
  41.26\%(0.45\%) &
  38.81\%(0.33\%) &
  37.01\%(0.27\%) &
  38.76\%(0.39\%) &
  34.83\%(0.44\%) \\
  \bottomrule
\end{tabular}%
}
\end{table*}
\section{Evaluations Beyond R+R } \label{sec:exp:beyond_RR}

In this section, we answer research questions RQ3 to RQ6 beyond R+R experiments and investigate the computational cost of different methods. Additional experiments such as empirical privacy measurements are in~\cref{app:empirical-privacy}.

\subsection{RQ3: Which Part of the Model to Fine-tune?}
We explore several fine-tuning strategies, drawing from Luo et al. \cite{luo2021scalable} and Cattan et al. \cite{cattan2022fine}. We apply these strategies to six distinct datasets: CIFAR-10, EuroSAT, ISIC 2018, Caltech 256, SUN397, and Oxford Pet. These datasets present a range of domain gaps with respect to the pre-training dataset. For this study, we use the Vit\_base\_patch16\_224  model, pre-trained on ImageNet, and fine-tuned it with 5 epochs. We present the results in \cref{tab:finetune_results}. To ensure a fair comparison, we use fine-tuning the whole model as a baseline. In addition to applying methods from the DPML literature, we experimented with alternatives. Instead of magnitude-based parameter selection, we test random parameter selection for 1\%, 2\%, and 10\% of the parameters (``Random Subset''). Also, rather than selecting parameters dispersed across different blocks of the ViT model, we tested randomly selecting 2, 3, or 6 blocks for fine-tuning, a strategy we call ``partial training.''

Results~(\cref{tab:finetune_results}) show that both alternative approaches sometimes match or surpass the performance of Luo et al.~\cite{luo2021scalable} for the first three datasets. Considering all 6 datasets, only fine-tuning the last layer achieves stable good performance. This suggests that fine-tuning only a subset of the model is the important variable, not the specific subset, or that the fine-tuning strategy is dataset-specific such as domain gap and data size per class.

\takeaway{Although several papers propose sophisticated strategies for DP fine-tuning models, we find that none of them consistently outperform alternatives across all datasets. It appears that only fine-tuning the last layer performs well across datasets.}

\subsection{RQ4: Can Different Methods be Combined?}
Can two or more methods from the selected papers be combined to provide further improvements? Given that the techniques are different in nature they are compatible and can in many cases be applied in combination (e.g., self-augmentation + changing order + ScaleNorm).

Since large batch sizes and high noise levels can be combined with all other improvements, we use this as the default setting for all experiments. We find that order switching (Sander et al.~\cite{sander2022tan}) and the techniques proposed by De et al.~\cite{de2022unlocking} can be combined and achieve 78.10\% test accuracy on CIFAR-10. By contrast, combining Mixed Ghost Clipping with order switching slightly decreased performance for order 3.

Combining ScaleNorm (Klause et al.~\cite{klause2022differentially}) with De et al.'s techniques or order switching did not significantly increase performance. Specifically, combining ScaleNorm with De et al.'s techniques achieved an average test accuracy of 77.43\%, which is lower than only applying De et al.'s techniques (77.79\%) and less than the 1\% performance boost claimed in the paper~\cite{klause2022differentially}. Similarly, when combining ScaleNorm with De et al.'s techniques and order switching the average test accuracy increased by only 0.09\% (smaller than the standard deviation across different runs).

\takeaway{Combining DPML methods, even orthogonal ones, often does not provide cumulative improvements. Some combinations of methods actually decrease performance. Notable exceptions include large batch sizes and augmentation multiplicity.}

\begin{table*}[!h]
\caption{Detail of 11 selected papers about their Generalizability and Reliability. Note that effect size is computed using our reproduced results of the baseline they used in their papers and their proposed methods. We measured the running time using one A100 GPU, running each method for 3 epochs on CIFAR-10. For the baseline of time comparison, DP-SGD from scratch took 90.61 seconds, while SGD from scratch took 10.62 seconds. For fine-tuning, DP-SGD took 571.16 seconds, compared to 388.76 seconds for SGD. The Time Overhead in the table is the slowdown factor relative to the corresponding SGD baseline in the same setting (i.e., method time divided baseline time).
}
\label{tab:claimed_results} 
\centering
\resizebox{1\textwidth}{!}{%
\begin{tabular}{l l  c c c  |c c c c c  |c|cc}
\toprule
Paper & Method(s) & \multicolumn{3}{c}{Generalizability} & \multicolumn{5}{|c|}{Reliability}  & Effect size & \multicolumn{2}{c}{Time Overhead}\\
& &Multi-Settings & Datasets & Multi-$\varepsilon$ & Open source & Param. Search Account & Multi-runs & Statistically significant & Ablation &  & vs SGD & vs DP-SGD\\
\midrule

Bu et al.~\cite{bu2022scalable} & Clipping techniques & $\times$ & \checkmark & \checkmark & \checkmark & $\times$ & $\times$ & N/A\tablefootnote{Because the authors do not report standard deviation, we cannot determine whether the improvement is larger than the standard deviation.} & $\times$  & 0.44 &1.76 &1.20 \tablefootnote{We use the latest version of Opacus so the time comparison may be different from reported in their paper.} \\
Bu et al. \cite{bu2022automatic} & Clipping techniques & \checkmark & \checkmark & \checkmark & \checkmark & $\times$ & \checkmark& $\times$ & \checkmark & -0.22 &2.17 &1.48 \\
Klause et al.~\cite{klause2022differentially} & Model architecture & $\times$ & \checkmark & $\times$ & $\times$ & $\times$ & $\times$ & N/A & \checkmark & 1.12 &9.57 &1.12 \\
Sander et al.~\cite{sander2022tan} & Model architecture & $\times$ & \checkmark & $\times$ & \checkmark & $\times$ & \checkmark & \checkmark & \checkmark & 5.28 &8.89 &1.04\\
Dormann et al. \cite{dormann2021not} & Hyperparameter tuning & $\times$ & \checkmark & $\times$ & \checkmark & $\times$ & \checkmark& \checkmark & \checkmark & 1.91 &8.53 &1.00 \tablefootnote{We use this method as the DP-SGD baseline as it is well-known.}\\
De et al.~\cite{de2022unlocking} &Augmentation multiplicity & \checkmark & \checkmark & \checkmark & \checkmark & \checkmark & \checkmark & \checkmark & \checkmark & 12.22 &93.90 &11.01\\
Bao et al.~\cite{bao2024dp} &Augmentation multiplicity & \checkmark & \checkmark & \checkmark & \checkmark & \checkmark & \checkmark & \checkmark & \checkmark & 13.62 &99.93 &11.71\\
Tram\`er and Boneh~\cite{tramer2020differentially}  & Feature selection & $\times$ & \checkmark & \checkmark & \checkmark & $\times$& \checkmark & \checkmark & \checkmark & 9.77 &1.05 &0.12 \tablefootnote{Because feature pre-processing of this method is not counted, the running time for this method is lower.} \\

Cattan et al. \cite{cattan2022fine} & Fine-tuning technique & N/A  & $\times$ & \checkmark & $\times$ & $\times$ & $\times$ & N/A & $\times$ & -21.14 &0.43 &0.29\\

Luo et al. \cite{luo2021scalable} & Fine-tuning technique & N/A  & \checkmark & \checkmark & $\times$ & $\times$ & $\times$ & N/A & \checkmark & 0.42 &11.84 &8.06 \tablefootnote{We implement this technique ourself so it is not be optimized for minimizing running time.}\\

Tang et al. \cite{tang2023differentially} & Fine-tuning technique & N/A  & \checkmark & \checkmark & \checkmark & \checkmark & \checkmark & \checkmark & \checkmark & 46.76 &93.16 &10.92 \\

\bottomrule
\end{tabular}%
}
\end{table*}

\subsection{RQ5: What are the Most Promising Methods?}
%
%

Our experiments show that while we could reproduce all selected papers, {\bf only seven out of eleven} papers including Dormann et al. \cite{dormann2021not}, Sander et al.~\cite{sander2022tan}, De et al.~\cite{de2022unlocking}, Bao et al. \cite{bao2024dp}, Luo et al. \cite{luo2021scalable}, Tram\`er and Boneh~\cite{tramer2020differentially} and Tang et al. \cite{tang2023differentially} reliably and consistently achieved their claimed performance improvements. The other four papers did not for various reasons, including not delivering consistent improvements outside of the narrow experimental setting in their paper. 

As a further demonstration of our proposed framework, we summarize its application to some of the evaluated techniques in~\cref{tab:summary}. We observe that (in this case) feature selection and augmentation multiplicity techniques achieve substantial improvements (i.e., large effect sizes) that are also statistical significance. By contrast, the clipping and fine-tuning techniques in this case do not achieve statistical significance (or large effect size). 


Further discussion of statistical power and number of runs using~\mbox{\cref{tab:summary}} for illustration is warranted. Reporting results for a few runs (sometimes a single run) is common practice in DPML research~\cite{bu2022scalable,klause2022differentially,luo2021scalable,bao2024dp,sander2022tan}. Accordingly, we used $n=3$ for the experiments in the table. Ideally, research should involve more extensive testing (e.g., $n=20$), but the high computational cost of DPML training makes this challenging for many researchers. We discuss the computational overhead of DPML in the next subsection. 

Reporting too few runs may result in an under-powered test, and being unable to establish whether the method provides notable improvements. However, as discussed in~\mbox{\cref{{sec:methodology:framework}}}, a large enough effect size can overcome a small number of runs. Power analysis on the results of~\mbox{\cref{tab:summary}} show that for~\mbox{\cite{tramer2020differentially,de2022unlocking}} the test has plenty of power due to the large effect sizes. By contrast, for the other two methods in the table, the effect size is too small. 

We reiterate that the goal of our R+R experiment is not to assign blame or cast any specific work in a negative light. Rather we seek to identify what methods and techniques work best and how to perform the evaluation of DPML to achieve high degrees of reproducibility (\cref{sec:reproducibility:guidelines}).

\begin{table}[tb]
\centering
\caption{Summary of statistical framework results. We use CIFAR-10 for all, except for Bu et al. \cite{bu2022automatic} where we used MNIST. We set run times $n=3$. We use Vit-base-patch16-224 for Bu et al. \cite{bu2022automatic} and Luo et al. \cite{luo2021scalable}, WideResNet-16-4 for De et al. \cite{de2022unlocking} and ScatterNet+CNN for Tram\`er and Boneh~\cite{tramer2020differentially}.}
\label{tab:summary}
\resizebox{0.9\columnwidth}{!}{%
\begin{tabular}{ccccc}
\toprule
Method               & Paper                             & t     & p-value & Effect size \\ \midrule
Clipping technique   & Bu et al. \cite{bu2022automatic}  & -0.43 & 0.71    & -0.22       \\
Feature selection         & Tram\`er and Boneh~\cite{tramer2020differentially} & 21.25 & {\bf 0.002} & 9.77  \\
Augmentation multiplicity & De et al.~\cite{de2022unlocking}                   & 22.01 & {\bf 0.002} & 12.22 \\
Fine-tuning technique & Luo et al.~\cite{luo2021scalable} & 0.49  & 0.67    & 0.42        \\ \bottomrule
\end{tabular}%
}
\end{table}
In the rest of this section, we discuss insights from investigations and highlight promising future research directions.

\paragraphbe{Model architecture, feature, and hyperparameter selection.}
Whenever possible comprehensive searches over model architecture ~\cite{bao2022importance,priyanshu2021efficient}, features~\cite{tramer2020differentially}, and hyperparameters~\cite{de2022unlocking,sander2022tan} should be performed as all of these factors play a pivotal role in DPML performance. Ideally, the privacy cost of hyperparameter searches should be accounted for, which will likely reduce the obtained performance. There is promising recent work in this direction such as\cite{papernot2021hyperparameter,ding2022revisiting,koskela2024practical}, but more research is necessary.

\paragraphbe{Large batch sizes.}
Larger batch sizes and higher noise levels provide consistently higher performance according to numerous studies~\cite{dormann2021not,de2022unlocking,sander2022tan} and our own empirical findings. However, large batch size brings a problem of high computational cost for hyperparameter tuning. To address this, techniques like Sander et al. \cite{sander2022tan} can be applied.

\paragraphbe{Clipping strategies.}
Although there is a plethora of papers exploring the use of clipping techniques to improve DPML, we find that such methods provide little improvement. Notably, several recent works \cite{xiao2023theory,koloskova2023revisiting,chen2020understanding} investigate the effect of clipping and how it may bias the learning process. 


\paragraphbe{Augmentation multiplicity.}
The augmentation multiplicity (self-augmentation) approaches of De et al.~\cite{de2022unlocking} and Bao et al. \cite{bao2024dp} appear to deliver consistent and significant improvements in model performance. However, these works only explore a small subset of possible augmentations, so a promising avenue for future research is to comprehensively study the potential benefits of various data augmentation techniques.

\paragraphbe{Architecture-specific methods.}%
Some methods such as changing the order of layers~\cite{sander2022tan} seem to provide improvements while others such as ScaleNorm~\cite{klause2022differentially} did not in our reproduction. Practitioners should be caution before adopting methods specifically tailored to an architecture as improvements may not be consistently obtained.

\paragraphbe{Fine-tuning methods.}
Fine-tuning a subset of a model's parameters with DP appears to be a viable strategy. However, no single method except only fine-tuning the last layer performs best across datasets and architecture in our experiments. Practitioners should attempt only fine-tuning the last layer and apply whatever method gives the best performance for their particular use cases rather than adopting whole cloth any of the methods that claim to provide SoTA results.

\paragraphbe{Pre-training and public data.}
Fine-tuning a pre-trained model with DP is a consistent way to achieve performance closer to the non-private setting than training from scratch. That said, the more similar the pre-training data and fine-tuning datasets are, the better the performance. When no suitable public dataset is available for pre-training, techniques such as Tram\`er and Boneh~\cite{tramer2020differentially} or Tang et al.~\cite{tang2023differentially} provide a viable alternative. However, our results suggest that using an unrelated public dataset for pre-training provides comparable results. 

\subsection{Computational Cost}\label{sec:overhead}
We evaluate computational cost by measuring the average GPU time per epoch for different methods over 3 epochs, all run on a single A100 GPU. Our benchmarks included training WRN-16-4 from scratch and fine-tuning the ViT model on CIFAR-10 using both SGD and DP-SGD. Results showed: Training from scratch with DP-SGD required 90.61 sec/epoch, whereas training with SGD took only 10.62 sec/epoch. For fine-tuning, DP-SGD took 571.16 sec/epoch, while SGD took 388.76 sec/epoch. We calculate the time overhead as a ratio of time per epoch to both SGD and DP-SGD. Results are shown in~\mbox{\cref{tab:claimed_results}}. 

We observe that more complex training strategies, such as those proposed by De et al. \mbox{\cite{de2022unlocking}}, Bao et al. \mbox{\cite{bao2024dp}}, Luo et al. \mbox{\cite{luo2021scalable}}, and Tang et al. \mbox{\cite{tang2023differentially}}, require significantly longer training times per epoch. On the other hand, techniques like feature selection \mbox{\cite{tramer2020differentially}} or fine-tuning fewer layers \mbox{\cite{cattan2022fine}} can substantially reduce computational time. 

\begin{table}[t!]
\centering
\caption{Proposed checklist for DPML.}
\label{tab:checklist}
\resizebox{0.875\columnwidth}{!}{%
\begin{tabular}{clc}
\toprule
\multicolumn{2}{c}{Checklist item}                                                      &  \\ \midrule
\multirow{5}{*}{Generalizability} & Evaluation in different settings                & \Square         \\
                                 
                                  & Evaluation with different datasets              & \Square        \\
                               
                                  & Evaluation with different model architectures   &\Square            \\
                                  & Evaluation for different privacy requirements   & \Square           \\
                                  & Evaluation of combination with other techniques & \Square           \\ \midrule
\multirow{5}{*}{Reliability}      & Code open source                                & \Square           \\
                                  & Results of multiple runs are reported           &\Square            \\
                                  & Improvement is statistically significant    & \Square            \\
                                  & Accounts for hyperparameter search            & \Square           \\ 
 & Includes ablation study   & \Square           \\ \bottomrule
\end{tabular}
}
\end{table}

\section{Towards Reproducibility \& Replicability}\label{sec:reproducibility:guidelines}

We distill our insights from our R+R experiment into a set of proposed guidelines and a checklist. We hope researchers who seek to evaluate new methods can follow these guidelines to maximize the chance of reproducibility.

\paragraphb{Criteria}
We propose to think of reproducibility and replicability along two separate axes: generalizability, and reliability.

\begin{itemize}[leftmargin=1.125em]
    \item {\bf\em Generalizability} assess whether the proposed method's benefits are likely to generalize outside of the narrow experimental setting demonstrated. For example, if a method was shown to provide improvements in multiple settings, varied datasets, and multiple privacy regimes, it is more likely to provide similar improvements in other contexts than a method only evaluated on a single task, dataset, and privacy budget.
\smallskip
    \item {\bf\em Reliability} assess the extent to which evaluation methodology suggests results reported are reliable, stable, and likely to reproduce. For instance, results from a single run are less reliable than those averaged over five independent runs. Additionally, reliability involves determining whether performance improvements are truly due to the proposed method, especially when combined techniques or unique settings might skew results. In such cases, the apparent enhancements in performance may stem from these ancillary factors rather than from the intrinsic merits of the proposed method.
\smallskip
    
\end{itemize}

\noindent The checklist is shown in~\cref{tab:checklist}.

\paragraphb{Selected papers}
We grade our 11 selected papers according to our checklist. Results are shown in \cref{tab:claimed_results}. We found that Dormann et al. \cite{dormann2021not},  De et al.\cite{de2022unlocking}, Bao et al. \cite{bao2024dp}, Tram\`er and Boneh~\cite{tramer2020differentially}, Luo et al. \cite{luo2021scalable}, Sander et al.\cite{sander2022tan} and Tang et al. \cite{tang2023differentially} performed well overall according to our two criteria. This is not the case for Klause et al.~\cite{klause2022differentially}, Bu et al.~\cite {bu2022scalable}, and other works. For example, Klause et al.~\cite{klause2022differentially} lack reliability (no open source and report results for only one run), Bu et al.~\cite{bu2022scalable} lack generalizability (only pre-trained tasks) and reliability (no ablation experiments), Bu et al. \cite{bu2022automatic} lacks reliability (report improvement without statistical difference) while Cattan et al. \cite{cattan2022fine} also lack reliability (no open source code, report results with one run).

\paragraphb{Checklist details}
We describe the items of the checklist, and their rationale, and illustrate their utility through examples in~\cref{app:checklist}.

\section{Conclusion and Future work}
We conducted a R+R investigation on 11 recent SoTA DPML techniques, which revealed significant variations in their reproducibility. We identified the inherent randomness of DPML as a challenge and proposed a statistical framework to deal with it. We distilled our insights into a set of comprehensive guidelines and a checklist to standardize future DPML research. Our investigation also uncovered open questions for future research, such as determining the optimal fine-tuning strategy with DP. The training convergence behavior of different DP methods is another possible direction for future work, and so is reproducibility of DPML methods for types of data other than images.

{
\section*{Acknowledgments}
This work was supported in part by the National Science Foundation under CNS-2055123 and CNS-1933208. Any opinions, findings, conclusions, or recommendations expressed in this material are those of the authors and do not necessarily reflect the views of the National Science Foundation.
}

\bibliographystyle{IEEEtran}
\bibliography{refs}

\appendices

\crefalias{section}{appendix}  
\crefalias{subsection}{appendix}     
\crefalias{subsubsection}{appendix}  

\nobalance

\section{Datasets} \label{app:dataset}
\paragraphb{\cifar{}}  
We use the \cifar{} dataset~\cite{krizhevsky2009learning}, which contains 60,000 images with 10 classes. We use 50,000 as the training set and 10,000 as the test set as following most papers do. Each image in \cifar{} has 3 RGB channels and its size is $32 \times 32$ pixels.

\paragraphb{MNIST} It~\cite{lecun1998mnist} contains 70,000 $28 \times 28 $ gray scale handwritten digit images. We use 60,000 for training and 10,000 for testing.

\paragraphb{Fashion-MNIST}
It \cite{xiao2017fashion} contains 70,000 $28 \times 28 $ grayscale images of clothing. We use 60,000 for training and 10,000 for testing.

\paragraphb{EuroSAT}
This dataset \cite{helber2019eurosat} contains Sentinel-2 satellite images with 10 classes. It has 27,000 $64 \times 64$ labeled color images. We use 21600 as the training set and 5400 as a test set.

\paragraphb{ISIC 2018}
We use task 3 of this 2018 year's dataset \footnote{\url{https://challenge.isic-archive.com}} published by the International Skin Imaging Collaboration (ISIC) for lesion classification challenges. It contains 10,015 images which we use 9,015 images for training and 1,000 for testing.

\paragraphb{PathMNIST}
This dataset is part of MedMNIST\cite{yang2021medmnist}. It contains 107,180 RGB images with 9 classes. The image size is $28 \times 28 $. We use 89,996 images as the training set and 7180 as the test set.

\paragraphb{Caltech 256} The Caltech 256 dataset~\cite{griffin2007caltech} is frequently utilized for image classification tasks, consisting of 30,607 RGB images across 257 diverse object categories. In our experiments, we allocated 80\% of these images for training purposes and the remaining 20\% for evaluation.

\paragraphb{SUN397} The Scene UNderstanding (SUN) \cite{xiao2016sun,xiao2010sun} dataset comprises 108,754 RGB images spanning 397 distinct classes. For our experimental framework, 80\% of these images were used for training, with the balance of 20\% reserved for testing.

\paragraphb{Oxford Pet} This dataset~\cite{parkhi2012cats} features 37 categories of cats and dogs, totaling 7,349 images. We used 3,680 images for the training set and 3,669 for the test set.

\section{Additional Experiments}

\subsection{Empirical Privacy Measurements} \label{app:empirical-privacy}

\begin{table}[h]
\centering
\caption{Membership Inference Attacks AUC for different methods for varying privacy budgets from $\varepsilon = 0.1$ to $\varepsilon = 8$.}
\label{tab:attack}
\resizebox{0.75\columnwidth}{!}{%
\begin{tabular}{cccccc}
\toprule
                          & $\varepsilon$ & P-Attack & S-Attack & R-Attack & C-Attack  \\ \midrule
\multirow{4}{*}{Basline}  & 0.1 & 0.50  & 0.50  & 0.49  & 0.50  \\
                          & 0.5 & 0.50  & 0.50  & 0.48  & 0.50  \\
                          & 1   & 0.49  & 0.50  & 0.48  & 0.50  \\
                          & 8   & 0.50  & 0.50  & 0.48  & 0.50  \\ \hline
\multirow{4}{*}{De et al. \cite{de2022unlocking}} & 0.1 & 0.50  & 0.50  & 0.50  & 0.50  \\
                          & 0.5 & 0.49  & 0.49  & 0.48  & 0.49  \\
                          & 1   & 0.50  & 0.50  & 0.48  & 0.50  \\
                          & 8   & 0.50  & 0.50  & 0.49  & 0.50  \\ \hline
\multirow{4}{*}{Bao et al. \cite{bao2024dp}}   & 0.1 & 0.50  & 0.50  & 0.50  & 0.50  \\
                          & 0.5 & 0.50  & 0.50  & 0.49  & 0.50  \\
                          & 1   & 0.50  & 0.50  & 0.48  & 0.50  \\
                          & 8   & 0.50  & 0.51  & 0.49  & 0.50 \\ \bottomrule
\end{tabular}%
}
\end{table}

Since DP is a worst-case notion, different methods providing the same DP guarantee could provide different empirical privacy, as measured by membership inference attacks~\mbox{\cite{shokri2017membership,ye2022enhanced,carlini2022membership}}. We use the popular Privacy Meter tool to run four different attacks.\footnote{\url{https://github.com/privacytrustlab/ml_privacy_meter}} The attacks are P-Attack (Population), R-Attack (Reference), S-Attack (Shadow Models) based on \mbox{\cite{ye2022enhanced}}, and C-Attack (Carlini et al.~\mbox{\cite{carlini2022membership}}). We report the Area Under the Curve (AUC) as a measure of the attack success rate.

We consider three methodologies: a baseline approach which involves training a WRN-16-4 network with vanilla-DP-SGD on CIFAR-10 from scratch, utilizing technique from De et al.~\mbox{\cite{de2022unlocking}}, and another leveraging method based on Bao et al.~\mbox{\cite{bao2024dp}}. Membership inference attacks require held-out data points from the training set, thus we limit the training dataset to 30,000 samples. The privacy budgets tested range from $\varepsilon = 0.1$ to $\varepsilon = 8$. Results are shown in ~\mbox{\cref{tab:attack}}. 

Despite the diversity in techniques, all methods maintain analogous levels of empirical privacy, achieving AUC close to 50\%—comparable to random guessing, even when using a relatively high privacy budget (i.e., $\varepsilon=8$).

\paragraphb{Empirical privacy and accuracy trade-off}
Results from~\mbox{\cref{tab:various_eps,tab:attack}} suggest that even loose privacy guarantees (e.g., $\varepsilon=8$) may offer meaningful protection. However, we caution that the attacks we perform are all based on the black-box setting, and that the conclusions may not hold with stronger attacks or in the white-box setting. We refer readers to the (growing) literature on privacy auditing (e.g.,\mbox{\cite{steinke2024privacy,jagielski2020auditing}}) for a more nuanced discussion.

\section{Checklist}\label{app:checklist}
In this section, we describe the rationale behind the items in our proposed checklist alongside examples.

\subsection{Generalizability}
\paragraphbe{Evaluation in different settings.}
If the proposed method can be used in multiple settings (e.g., train from scratch and pre-trained), it should be evaluated in different settings.
\begin{itemize}[leftmargin=1.25em]
    \item {\em Rationale}: A method could provide a substantial improvement in one setting but no improvement in another setting. 
    \item  {\em Examples}: De et al. \cite{de2022unlocking} advocate for self-augmentations, showcasing notable performance in both from-scratch training and fine-tuning, a finding corroborated by our reproduction experiments. 
\end{itemize}

\paragraphbe{Evaluation with different datasets.}
Methods should be evaluated with multiple different domain datasets to uncover whether improvements persists.
\begin{itemize}[leftmargin=1.25em]
    \item {\em Rationale}: Some methods may only provide improvements for some datasets because different datasets involve tasks of varying difficulty. For example, a method may work only on specific datasets, or datasets with few classes or few input features. Moreover, when considering pre-trained models, a method may appear to work particularly well because the pre-trained model was trained on data similar to the target dataset. 
    \item  {\em Examples}: Some papers only report results for few datasets like CIFAR-10 and CIFAR-100. When applying their methods to datasets from a different domain such as EuroSAT, we did not observe their claimed improvements.
\end{itemize}

\paragraphbe{Evaluation with different architectures.}
Methods should be evaluated with different model architectures, if applicable.
\begin{itemize}[leftmargin=1.25em]
    \item {\em Rationale}: Some methods may only provide improvements with specific model architectures because of the nature of the method or task. Moreover, performance on a task can greatly differ from one architecture to another. For example, a method may perform well using a model architecture with relatively few parameters because that architecture may be well-suited for the considered tasks. For different tasks, however, the method may falter if such tasks require models with much larger parameter counts.
    \item  {\em Examples}: Some papers only evaluate their methods on a particular model such as WRN-28-10 model. When applying their method to VIT model, we are not able to reproduce their claimed performance.
\end{itemize}

\paragraphbe{Evaluation for different privacy requirements.}
Methods should be evaluated in different privacy regimes, that is with different range of values for $\varepsilon$ (and $\delta$ if applicable). The privacy parameters range considered should be appropriate for the given setting. For example, pre-trained models fine-tuned with large datasets may tolerate much lower $\varepsilon$ values than models trained from scratch on small datasets.
\begin{itemize}[leftmargin=1.25em]
    \item {\em Rationale}: Improvements provided from a method may not be uniform across all privacy regimes. Typically the less stringent the privacy requirement the less improvement there is, in part because DPML performance is closer to the non-private setting than for more stringent requirements. 
    \item  {\em Examples}: Tram\`er and Boneh~\cite{tramer2020differentially} present their method's results for low privacy budgets, specifically values smaller than 3. In our reproduction of their experiments with a larger privacy budget, such as 8, we observed that the performance of their proposed method plateaued as the privacy budget increased.
\end{itemize}

\paragraphbe{Evaluation of combination with other techniques.}
Authors should evaluate or discuss whether their proposed methods can be combined with other methods. 
\begin{itemize}[leftmargin=1.25em]
    \item {\em Rationale}: Some combinations of seemingly orthogonal methods actually decrease performance. 
    \item  {\em Examples}: Combining the methods of De et al. \cite{de2022unlocking} with Sander et al. \cite{sander2022tan} yields improved performance. However, merging Klause et al. \cite{klause2022differentially} with Sander et al. \cite{sander2022tan} leads to a decrease in performance.
\end{itemize}

\subsection{Reliability}

\paragraphbe{Code open sourcing}
Authors should open-source their code to facilitate reproducibility.
\begin{itemize}[leftmargin=1.25em]
    \item {\em Rationale}: Differences in the implementation of the same technique or the use of different codebases can yield significant differences. Open-sourcing code is a straightforward way to mitigate such concerns.
\end{itemize}

\paragraphbe{Results of multiple runs are reported.}
Our experiments on the randomness of DPML show that the variability of DPML is significant (\cref{sec:random_seeds}). Reporting the result of multiple runs is a way to mitigate this problem.
\begin{itemize}[leftmargin=1.25em]
    \item {\em Rationale}: Providing both an aggregate measure and a measure of variability across runs (e.g., mean and std) facilitates scientifically valid comparisons. The number of runs performed should be appropriate for the setting and privacy budget.
    \item  {\em Examples}: Some papers do not present results from multiple runs, making it challenging to discern if the performance improvement is due to their proposed method or due to randomness.
\end{itemize}

\paragraphbe{Improvement is statistically significant}
The proposed techniques should show statistically significant improvement beyond baselines using our proposed framework.
\begin{itemize}[leftmargin=1.25em]
    \item {\em Rationale}: If the performance of two methods being compared is similar, then we cannot conclusively determine which method (if any) is superior.
    \item {\em Examples}: The performance boost from some papers is not statistically significant. 
\end{itemize}

\paragraphbe{Account for hyperparameter search.}
The cost and benefit of hyperparameter search need to be taken into account.
\begin{itemize}[leftmargin=1.25em]
    \item {\em Rationale}: To avoid unfair comparisons any hyperparameter search must be accounted for. Ideally, separate validation and test sets should be used and the privacy cost of the search should be reported.
    \item {\em Examples}: None of our selected papers pay the privacy budget to hyperparameter search. We also find that only De et al.\cite{de2022unlocking} and Tang et al. \cite{tang2023differentially} have validation set for hyperparameter tuning.
    
\end{itemize}


\paragraphbe{Evaluation includes ablation experiments.}
Authors should ensure that improvements observed can be attributed to the proposed methods, e.g., through the use of an ablation study or experimental methodology to exclude other factors. 
\begin{itemize}[leftmargin=1.25em]
    \item {\em Rationale}: Some papers combine multiple methods without independent evaluation, or otherwise evaluate their approach in a way that measured improvements cannot be conclusively tied back to the proposed technique. 
    \item {\em Examples}: Some papers do not provide a comprehensive ablation study. For example, when using a more complex pre-trained model than prior work to achieve new SoTA results, observed improvements could be due to a pre-trained model, the proposed method, or both.
\end{itemize}

\end{document}